\documentclass{article}

\usepackage[preprint]{neurips_2023}
\usepackage[dvipsnames]{xcolor}
\usepackage[utf8]{inputenc} 
\usepackage[T1]{fontenc}    
\usepackage{url}            
\usepackage{booktabs}       
\usepackage{amsfonts}       
\usepackage{nicefrac}       
\usepackage{microtype}      
\usepackage{multirow}
\usepackage{graphicx}
\usepackage{pythonhighlight}
\usepackage{bmpsize}
\usepackage{soul}
\usepackage{colortbl}

\usepackage{amsmath}
\usepackage{amssymb}
\usepackage{mathtools}
\usepackage{amsthm}
\usepackage{pifont}
\usepackage{subcaption}

\usepackage[colorlinks,citecolor=mydarkblue,urlcolor=mydarkblue,linkcolor=mydarkblue]{hyperref}

\usepackage[capitalize,noabbrev]{cleveref}

\theoremstyle{plain}
\newtheorem{theorem}{Theorem}[section]

\theoremstyle{definition}

\theoremstyle{remark}

\definecolor{forestgreen}{rgb}{0.0, 0.5, 0.0}
\definecolor{mydarkblue}{rgb}{0, 0, 0.5}

\newcommand{\deltavalup}[2]{$#1$\color{ForestGreen} $\uparrow$\textsubscript{$#2$}}

\DeclareMathAlphabet{\pazocal}{OMS}{zplm}{m}{n}

\newcommand{\by}{\bar y}

\newcommand{\pir}{\pi_{\mathrm{ref}}}

\title{Monte Carlo Tree Search Boosts Reasoning via Iterative Preference Learning}

\author{%
  Yuxi Xie$^{1}$\footnotemark[1]\thanks{Correspondence to: Yuxi Xie (\href{xieyuxi@u.nus.edu}{xieyuxi@u.nus.edu}) and Anirudh Goyal (\href{anirudhgoyal9119@gmail.com}{anirudhgoyal9119@gmail.com}).} \quad Anirudh Goyal \quad Wenyue Zheng$^{1}$ \quad Min-Yen Kan$^{1}$ \\
  \textbf{Timothy Lillicrap}$^{2}$ \quad  \textbf{Kenji Kawaguchi}$^{1}$ \quad \textbf{Michael Shieh}$^{1}$ \\
  $^1$ National University of Singapore \\
  $^2$ Google DeepMind \\
}

\begin{document}

\maketitle

\begin{abstract}
    We introduce an approach aimed at enhancing the reasoning capabilities of Large Language Models (LLMs) through an iterative preference learning process inspired by the successful strategy employed by AlphaZero. Our work leverages Monte Carlo Tree Search (MCTS) to iteratively collect preference data, utilizing its look-ahead ability to break down instance-level rewards into more granular step-level signals. To enhance consistency in intermediate steps, we combine outcome validation and stepwise self-evaluation, continually updating the quality assessment of newly generated data. The proposed algorithm employs Direct Preference Optimization (DPO) to update the LLM policy using this newly generated step-level preference data. Theoretical analysis reveals the importance of using on-policy sampled data for successful self-improving. Extensive evaluations on various arithmetic and commonsense reasoning tasks demonstrate remarkable performance improvements over existing models. For instance, our approach outperforms the Mistral-7B Supervised Fine-Tuning (SFT) baseline on GSM8K, MATH, and ARC-C, with substantial increases in accuracy to $81.8\%$ (+$5.9\%$), $34.7\%$ (+$5.8\%$), and $76.4\%$ (+$15.8\%$), respectively. Additionally, our research delves into the training and inference compute tradeoff, providing insights into how our method effectively maximizes performance gains. Our code is publicly available at \href{https://github.com/YuxiXie/MCTS-DPO}{https://github.com/YuxiXie/MCTS-DPO}.
\end{abstract}

\section{Introduction}\label{sec:intro}
Development of Large Language Models (LLMs), has seen a pivotal shift towards aligning these models more closely with human values and preferences~\citep{DBLP:journals/corr/abs-2009-01325, DBLP:conf/nips/Ouyang0JAWMZASR22, DBLP:journals/corr/abs-2204-05862}. A critical aspect of this process involves the utilization of preference data. There are two prevailing methodologies for incorporating this data: the first entails the construction of a reward model based on preferences, which is then integrated into a Reinforcement Learning (RL) framework to update the policy~\citep{DBLP:conf/nips/ChristianoLBMLA17,bai2022constitutional}; the second, more stable and scalable method, directly applies preferences to update the model's policy~\citep{DBLP:journals/corr/abs-2305-18290}.

In this context, the concept of ``iterative'' development is a key, especially when contrasted with the conventional Reinforcement Learning from Human Feedback (RLHF) paradigm~\citep{DBLP:conf/nips/ChristianoLBMLA17,DBLP:journals/corr/abs-2009-01325, DBLP:conf/nips/Ouyang0JAWMZASR22, DBLP:journals/corr/abs-2204-05862}, where the reward model is often trained offline and remains static. An iterative approach proposes a dynamic and continuous refinement process~\citep{DBLP:conf/nips/ZelikmanWMG22, DBLP:journals/corr/abs-2308-08998, DBLP:conf/emnlp/0001GHW00023, yuan2024self}. It involves a cycle that begins with the current policy, progresses through the collection and analysis of data to generate new preference data, and uses this data to update the policy. This approach underlines the importance of ongoing adaptation in LLMs, highlighting the potential for these models to become more attuned to the complexities of human decision-making and reasoning.

\begin{figure*}
  \centering
  \includegraphics[width=.72\textwidth]{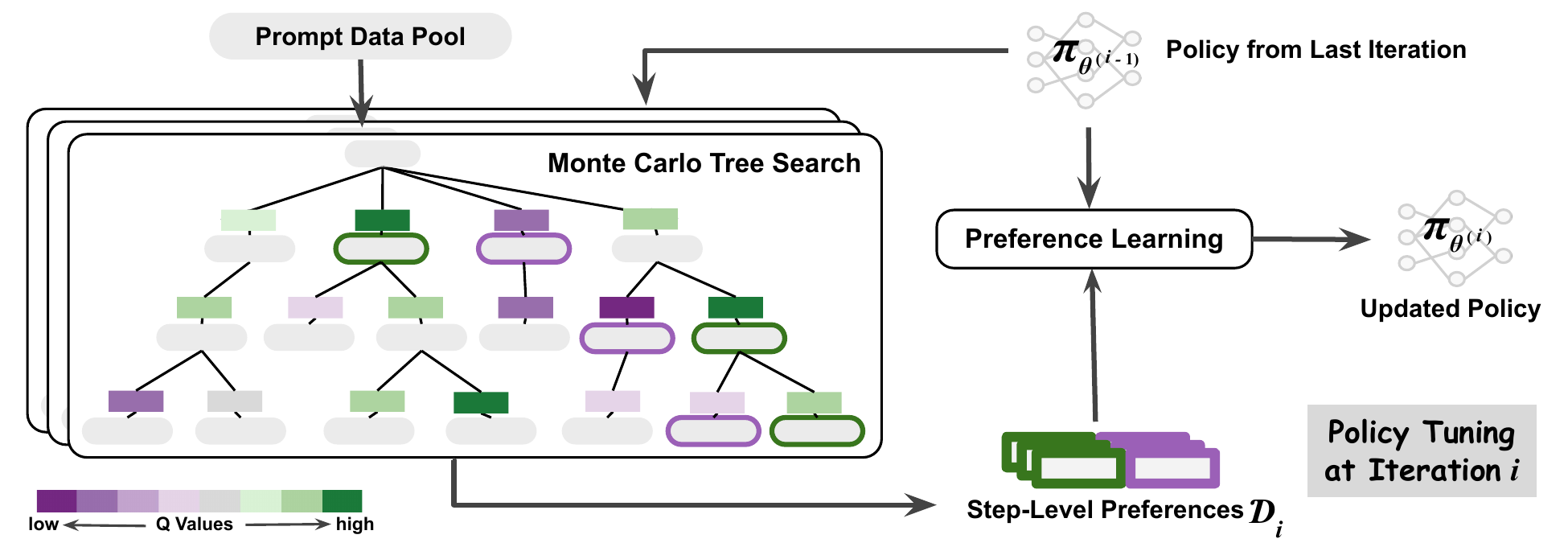}
    \hfill
    \includegraphics[width=.27\textwidth]{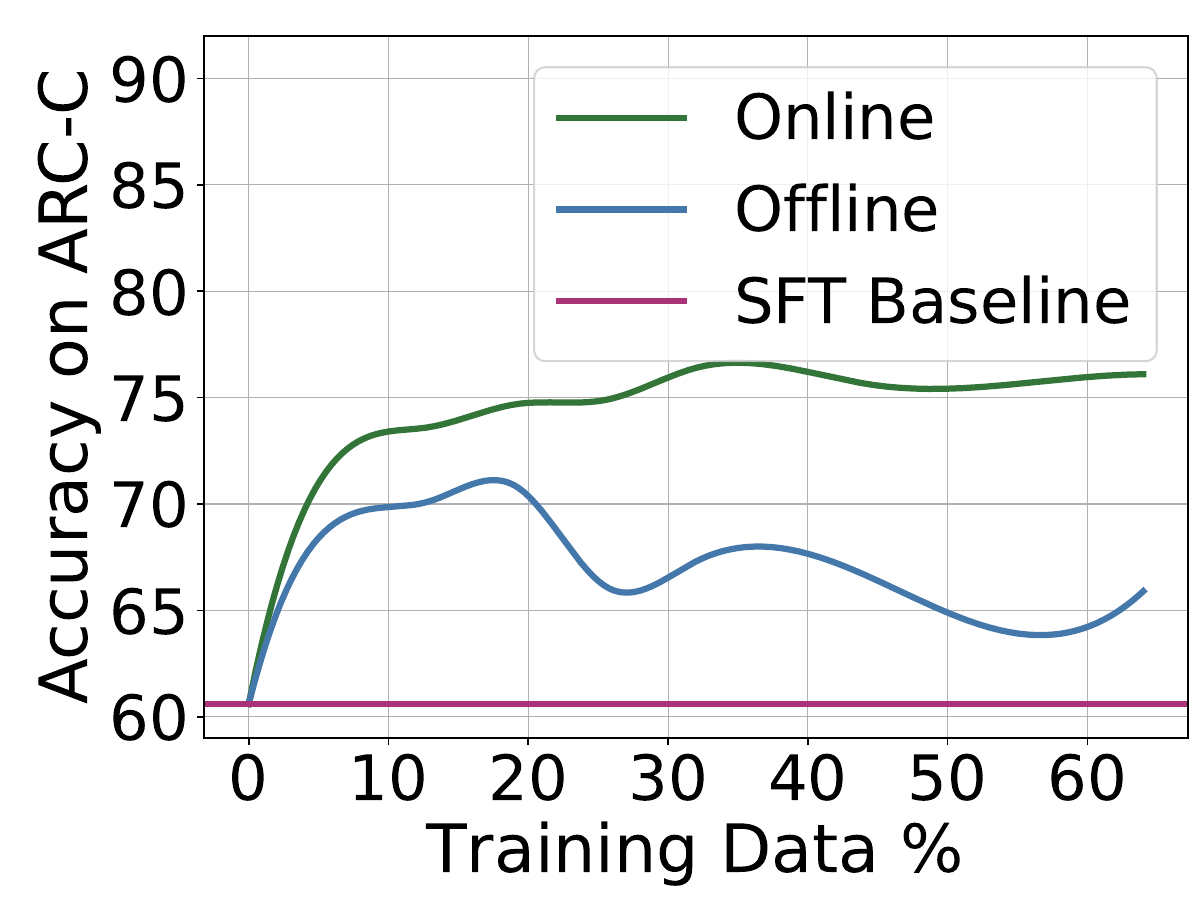}
    \vskip -0.1in
  \caption{Monte Carlo Tree Search (MCTS) boosts model performance via iterative preference learning. Each iteration of our framework (on the left) consists of two stages: MCTS to collect step-level preferences and preference learning to update the policy. Specifically, we use action values $Q$ estimated by MCTS to assign the preferences, where steps of higher and lower $Q$ values will be labeled as \textcolor[RGB]{56,118,29}{positive} and \textcolor[RGB]{155,97,183}{negative} data, respectively. The scale of $Q$ is visualized in the colormap. We show the advantage of the online manner in our iterative learning framework using the validation accuracy curves as training progresses on the right. The performance of ARC-C validation illustrates the effectiveness and efficiency of our proposed method compared to its offline variant.
  }
  \label{fig:framework}
  \vskip -0.2in
\end{figure*}

A compelling illustration of the success of such an iterative approach can be seen in the case of AlphaZero~\citep{DBLP:journals/corr/abs-1712-01815} for its superhuman performance across various domains, which combines the strengths of neural networks, RL techniques, and Monte Carlo Tree Search (MCTS)~\citep{coulom2006efficient, kocsis2006bandit}. The integration of MCTS as a \textit{policy improvement operator}  that transforms the current policy into an improved policy ~\citep{grill2020monte}. The effectiveness of AlphaZero underscores the potential of combining these advanced techniques in LLMs. By integrating MCTS into the iterative process of policy development, it is plausible to achieve significant strides in LLMs, particularly in the realm of reasoning and decision-making aligned with human-like preferences~\citep{DBLP:conf/acl/ZhuWZZ0GZY23, DBLP:conf/emnlp/HaoGMHWWH23}.

The integration of MCTS in collecting preference data to improve the current policy iteratively is nuanced and demands careful consideration. One primary challenge lies in determining the appropriate granularity for applying MCTS. Conventionally, preference data is collected at the instance level. The instance-level approach employs sparse supervision, which can lose important information and may not optimally leverage the potential of MCTS in improving the LLMs~\citep{DBLP:journals/corr/abs-2306-01693}. Another challenge is the reliance of MCTS on a critic or a learned reward function. This function is crucial for providing meaningful feedback on different rollouts generated by MCTS, thus guiding the policy improvement process~\citep{DBLP:journals/corr/abs-2309-15028}.

Addressing this granularity issue, evidence from LLM research indicates the superiority of process-level or stepwise evaluations over instance-level ones~\citep{DBLP:journals/corr/abs-2305-20050, li2023making, DBLP:journals/corr/abs-2305-00633, DBLP:journals/corr/abs-2305-10601}. Our approach utilizes MCTS rollouts for step-level guidance, aligning with a more granular application of MCTS. Moreover, we employ self-evaluation, where the model assesses its outputs, fostering a more efficient policy improvement pipeline by acting as both policy and critic~\citep{DBLP:journals/corr/abs-2207-05221, DBLP:journals/corr/abs-2305-00633}. This method streamlines the process and ensures more cohesive policy updates, aligning with the iterative nature of policy enhancement and potentially leading to more robust and aligned LLMs.

To summarize, we propose an algorithm based on Monte Carlo Tree Search (MCTS) that breaks down the instance-level preference signals into step-level. MCTS allows us to use the current LLM policy to generate preference data instead of a predetermined set of human preference data, enabling the LLM to receive real-time training signals. During training, we generate sequences of text on the fly and label the preference via MCTS based on feedback from self-evaluation (Figure~\ref{fig:framework}). To update the LLM policy using the preference data, we use Direct Preference Optimization (DPO)~\citep{DBLP:journals/corr/abs-2305-18290}. 
We extensively evaluate the proposed approach on various arithmetic and commonsense reasoning tasks and observe significant performance improvements. For instance, the proposed approach outperforms the Mistral-7B SFT baseline by $81.8\%$ (+$5.9\%$), $34.7\%$ (+$5.8\%$), and $76.4\%$ (+$15.8\%$) on GSM8K, MATH, and SciQ, respectively. Further analysis of the training and test compute tradeoff shows that our method can effectively push the performance gains in a more efficient way compared to sampling-only approaches.

\section{MCTS-Enhanced Iterative Preference Learning}\label{sec:method}
In this paper, we introduce an approach for improving LLM reasoning, centered around an iterative preference learning process. The proposed method begins with an initial policy $\pi_{\theta^{(0)}}$, and a dataset of prompts $\mathcal{D}_{\mathcal{P}}$. Each iteration $i$ involves selecting a batch of prompts from $\mathcal{D}_{\mathcal{P}}$, from which the model, guided by its current policy $\pi_{\theta^{(i-1)}}$, generates potential responses for each prompt. We then apply a set of dynamically evolving reward criteria to extract preference data $\mathcal{D}_i$ from these responses. The model's policy is subsequently tuned using this preference data, leading to an updated policy $\pi_{\theta^{(i)}}$, for the next iteration. This cycle of sampling, response generation, preference extraction, and policy tuning is repeated, allowing for continuous self-improvement and alignment with evolving preferences. In addressing the critical aspects of this methodology, two key challenges emerge: the effective collection of preference data and the process of updating the policy post-collection.

We draw upon the concept that MCTS can act as an approximate policy improvement operator, transforming the current policy into an improved one. Our work leverages MCTS to iteratively collect preference data, utilizing its look-ahead ability to break down instance-level rewards into more granular step-level signals. To enhance consistency in intermediate steps, we incorporate stepwise self-evaluation, continually updating the quality assessment of newly generated data. This process, as depicted in Figure~\ref{fig:framework}, enables MCTS to balance quality exploitation and diversity exploration during preference data sampling at each iteration. Detailed in section \ref{sec:mcts}, our approach utilizes MCTS for step-level preference data collection. Once this data is collected, the policy is updated using DPO, as outlined in section \ref{sec:target}. Our method can be viewed as an online version of DPO, where the updated policy is iteratively employed to collect preferences via MCTS. Our methodology, thus, not only addresses the challenges in preference data collection and policy updating but also introduces a dynamic, iterative framework that significantly enhances LLM reasoning.

\begin{algorithm*}[tb]
\small
   \caption{\textbf{. MCTS-Enhanced Iterative Preference Learning.} Given an initial policy $\pi_{\theta^{(0)}}=\pi_{\mathrm{sft}}$, our algorithm iteratively conducts step-level preference data sampling via MCTS and preference learning via DPO to update the policy.}
   \label{alg:example}
\begin{algorithmic}
   \STATE {\bfseries Input:} $\mathcal{D}_{\mathcal{P}}$: prompt dataset; $q(\cdot\mid x)$: MCTS sampling strategy that constructs a tree-structured set of possible responses given a prompt $x$, where $q_{\pi}$ represents that the strategy is based on the policy $\pi$ for both response generation and self-evaluation; $\ell_i(x, y_w, y_l;\theta)$: loss function of preference learning at the $i$-th iteration, where the corresponding sampling policy is $\pi^{(i)}$; $M$: number of iterations; $B$: number of samples per iteration; $T$: average number of steps per sample
   \STATE Train $\pi_{\theta}$ on $\mathcal{D}_{\mathcal{P}}$ using step-level preference learning.
   \FOR{$i=1$ {\bfseries to} $M$}
   \STATE $\pi^{(i)} \leftarrow \pi_{\theta} \leftarrow \pi_{\theta^{(i-1)}}$ 
   \STATE Sample a batch of $B$ samples from $\mathcal{D}_{\mathcal{P}}$ as $\mathcal{D}_{\mathcal{P}}^{(i)}$.
   \STATE \COMMENT{MCTS for Step-Level Preference Data Collection}
   \STATE For each $x\in \mathcal{D}_{\mathcal{P}}^{(i)}$, elicit a search tree of depth $T$ via $q_{\pi_{\theta}}(\cdot\mid x)$.
   \STATE Collect a batch of preferences $\mathcal{D}_i = \{$ $\{(x^j, y_l^{(j, t)}, y_l^{(j, t)})|_{t=1}^T\}|_{j=1}^B $ s.t. $x^j\sim \mathcal{D}_{\mathcal{P}}^{(i)}, y_w^{(j, t)} \neq y_w^{(j, t)} \sim q_{\pi_{\theta}}(\cdot\mid x^j)$ $\}$, where $y_w^{(j,t)}$ and $y_l^{(j,t)}$ is the nodes at depth $t$, with the highest and lowest $Q$ values, respectively, among all the children nodes of their parent node. 
   \STATE \COMMENT{Preference Learning for Policy Improvement}
   \STATE Optimize $\theta$ by minimizing $J(\theta) = \mathbb{E}_{(x, y_w, y_l)\sim \mathcal{D}_i} \ell_i(x, y_w, y_l; \theta)$.
   \STATE Obtain the updated policy $\pi_{\theta^{(i)}}$
   \ENDFOR
   \STATE $\pi_{\theta} \leftarrow \pi_{\theta^{(M)}}$
   \STATE {\bfseries Output:} Policy $\pi_{\theta}$
\end{algorithmic}
\end{algorithm*}

\subsection{MCTS for Step-Level Preference}\label{sec:mcts}

To transform instance-level rewards into granular, step-level signals, we dissect the reasoning process into discrete steps, each represented by a token sequence. We define the state at step $t$, $s_t$, as the prefix of the reasoning chain, with the addition of a new reasoning step $a$ transitioning the state to $s_{t+1}$, where $s_{t+1}$ is the concatenation of $s_t$ and $a$. Utilizing the model's current policy $\pi_{\theta}$, we sample candidate steps from its probability distribution $\pi_{\theta}(a\mid x, s_t)$\footnote{For tasks (\textit{e.g.}, MATH) where the initial policy performs poorly, we also include the ground-truth reasoning steps for training. We detail the step definition for different tasks with examples in Appendices~\ref{app:implementation} and \ref{app:analysis}.}, with $x$ being the task's input prompt. MCTS serves as an approximate policy improvement operator by leveraging its look-ahead capability to predict the expected future reward. This prediction is refined through stepwise self-evaluation~\citep{DBLP:journals/corr/abs-2207-05221, DBLP:journals/corr/abs-2305-00633}, enhancing process consistency and decision accuracy. The tree-structured search supports a balance between exploring diverse possibilities and exploiting promising paths, essential for navigating the vast search space in LLM reasoning.

The MCTS process begins from a root node, $s_0$, as the sentence start or incomplete response, and unfolds in three iterative stages: selection, expansion, and backup, which we detail further.

\paragraph{Select.}
The objective of this phase is to identify nodes that balance search quality and computational efficiency. The selection is guided by two key variables: $Q(s_t, a)$, the value of taking action $a$ in state $s_t$, and $N(s_t)$, the visitation frequency of state $s_t$. These variables are crucial for updating the search strategy, as explained in the backup section. To navigate the trade-off between exploring new nodes and exploiting visited ones, we employ the Predictor + Upper Confidence bounds applied to Trees (PUCT)~\citep{rosin2011multi}. At node $s_t$, the choice of the subsequent node follows the formula:

\begin{equation}
    \label{eq:puct}
    \begin{aligned}
        {s_{t+1}}^* &= \arg\max_{s_t}\Bigl[Q(s_t, a) + c_{\mathrm{puct}}\cdot p(a\mid s_t)\frac{\sqrt{N(s_t)}}{1 + N(s_{t+1})}\Bigl]
    \end{aligned}
\end{equation}

where $p(a\mid s_t) = \pi_{\theta}(a\mid x, s_t)/|a|^{\lambda}$ denotes the policy $\pi_{\theta}$'s probability distribution for generating a step $a$, adjusted by a $\lambda$-weighted length penalty to prevent overly long reasoning chains.

\paragraph{Expand.}
Expansion occurs at a leaf node during the selection process to integrate new nodes and assess rewards. The reward $r(s_t, a)$ for executing step $a$ in state $s_t$ is quantified by the reward difference between states $R(s_t)$ and $R(s_{t+1})$, highlighting the advantage of action $a$ at $s_t$. As defined in Eq.~(\ref{eq:reward}), reward computation merges outcome correctness $\mathcal{O}$ with self-evaluation $\mathcal{C}$. We assign the outcome correctness to be $1$, $-1$, and $0$ for correct terminal, incorrect terminal, and unfinished intermediate states, respectively. Following \citet{DBLP:journals/corr/abs-2305-00633}, we define self-evaluation as Eq.~(\ref{eq:selfeval}), where $\mathrm{A}$ denotes the confidence score in token-level probability for the option indicating correctness\footnote{We show an example of evaluation prompt in Table~\ref{tab:eval-prompt}.}. Future rewards are anticipated by simulating upcoming scenarios through roll-outs, following the selection and expansion process until reaching a terminal state\footnote{The terminal state is reached when the whole response is complete or exceeds the maximum length.}. 
\begin{equation}
    \label{eq:reward}
    R(s_t) = \mathcal{O}(s_t) + \mathcal{C}(s_t)
\end{equation}
\begin{equation}
    \label{eq:selfeval}
    \mathcal{C}(s_t) = \pi_{\theta}(\mathrm{A}\mid {\mathrm{prompt}}_{\mathrm{eval}}, x, s_t)
\end{equation}

\paragraph{Backup.}
Once a terminal state is reached, we carry out a bottom-up update from the terminal node back to the root. We update the visit count $N$, the state value $V$, and the transition value $Q$: 
\begin{equation}
    \label{eq:Q}
    Q(s_t, a) \leftarrow r(s_t, a) + \gamma V(s_{t+1})
\end{equation}
\begin{equation}
    \label{eq:V}
    V(s_t) \leftarrow \sum_{a}N(s_{t+1})Q(s_t, a) / \sum_{a}N(s_{t+1})
\end{equation}
\begin{equation}
    \label{eq:N}
    N(s_t) \leftarrow N(s_t) + 1
\end{equation}
where $\gamma$ is the discount for future state values.

For each step in the response generation, we conduct $K$ iterations of MCTS to construct the search tree while updating $Q$ values and visit counts $N$. To balance the diversity, quality, and efficiency of the tree construction, we initialize the search breadth as $b_1$ and anneal it to be a smaller $b_2 < b_1$ for the subsequent steps. We use the result $Q$ value corresponding to each candidate step to label its preference, where higher $Q$ values indicate preferred next steps. For a result search tree of depth $T$, we obtain $T$ pairs of step-level preference data. Specifically, we select the candidate steps of highest and lowest $Q$ values as positive and negative samples at each tree depth, respectively. The parent node selected at each tree depth has the highest value calculated by multiplying its visit count and the range of its children nodes' visit counts, indicating both the quality and diversity of the generations. 

\subsection{Iterative Preference Learning}\label{sec:target}
Given the step-level preferences collected via MCTS, we tune the policy via DPO~\citep{DBLP:journals/corr/abs-2305-18290}. Considering the noise in the preference labels determined by $Q$ values, we employ the conservative version of DPO~\citep{eric2023dponote} and use the visit counts simulated in MCTS to apply adaptive label smoothing on each preference pair. Using the shorthand $h_{\pi_{\theta}}^{y_w, y_l} = \log  \frac{\pi_{\theta}(y_w\mid x)}{\pir(y_w\mid x)} - \log  \frac{\pi_{\theta}(y_l\mid x)}{\pir(y_l\mid x)}$, at the $i$-th iteration, given a batch of preference data $\mathcal{D}_i$ sampled with the latest policy $\pi_{\theta^{(i-1)}}$, we denote the policy objective $\ell_i(\theta)$ as follows:
\begin{equation}
    \label{eq:loss}
    \begin{aligned}
        \ell_{i}(\theta) =&  -\mathbb{E}_{(x,y_w,y_l)\sim\mathcal{D}_i} \Bigr[(1 - \alpha_{x, y_w, y_l})\log \sigma ( \beta \left. h_{\pi_{\theta}}^{y_w, y_l} \right) + &\alpha_{x, y_w, y_l} \log \sigma ( - \beta h_{\pi_{\theta}}^{y_w, y_l} ) \Bigr]
    \end{aligned}
\end{equation}
where $y_w$ and $y_l$ represent the step-level preferred and dispreferred responses, respectively, and the hyperparameter $\beta$ scales the KL constraint. Here, $\alpha_{x, y_w, y_l}$ is a label smoothing variable calculated using the visit counts at the corresponding states of the preference data $y_w$, $y_l$ in the search tree:
\begin{equation}
    \label{eq:label-smooth}
    \alpha_{x, y_w, y_l} = \frac{1}{N(x, y_w) / N(x, y_l) + 1}
\end{equation}
where $N(x, y_w)$ and $N(x, y_l)$ represent the states taking the actions of generating $y_w$ and $y_l$, respectively, from their previous state as input $x$. 

After optimization, we obtain the updated policy $\pi_{\theta^{(i)}}$ and repeat the data collection process in Section~\ref{sec:mcts} to iteratively update the LLM policy. We outline the full algorithm of our MCTS-enhanced Iterative Preference Learning in Algorithm~\ref{alg:example}.

\section{Theoretical Analysis}~\label{sec:theory}
Our approach can be viewed as an online version of DPO, where we iteratively use the updated policy to sample preferences via MCTS. In this section, we provide theoretical analysis to interpret the advantages of our online learning framework compared to the conventional alignment techniques that critically depend on offline preference data. We review the typical RLHF and DPO paradigms in Appendix~\ref{app:online-dpo}.

We now consider the following abstract formulation for clean theoretical insights to analyze our online setting of preference learning. 
Given a prompt $x$, there exist $n$ possible suboptimal responses $\{\by_1,\dots,\by_n\}=Y$ and an optimal outcome $y^{*}$. 
As specified in Equation~\ref{eq:loss}, at the $i$-th iteration, a pair of responses $(y,{y^{\prime}})$ are sampled from some sampling policy $\pi^{(i)}$ without replacement so that $y\neq {y^{\prime}}$ as $y \sim\pi^{(i)}(\cdot\mid x)$ and ${y^{\prime}} \sim\pi^{(i)}(\cdot\mid x,y)$.
Then, these are labeled to be $y_w$ and $y_l$ according to the preference. Define $\Theta$ be a set of all global optimizers of the preference loss for all $M$ iterations, \textit{i.e.}, for any $\theta \in \Theta$, $\ell_{i}(\theta) =0$ for all $i\in \{1,2,\cdots,M\}$. Similarly, let ${\theta^{(i)}}$ be a parameter vector such that $\ell_j(\theta^{(i)})= 0$ for all $j\in \{1,2,\cdots,i-1\}$ for $i \ge 1$ whereas $\theta^{(0)}$ is the initial parameter vector.

This abstract formulation covers both the offline and online settings. The offline setting in previous works is obtained by setting $\pi^{(i)} =\pi$ for some fixed distribution $\pi$. The online setting is obtained by setting $\pi^{(i)}=\pi_{\theta^{(i-1)}}$ where $\pi_{\theta^{(i-1)}}$ is the latest policy at beginning of the $i$-th iteration.

The following theorem shows that the offline setting can fail with high probability if the sampling policy $\pi^{(i)}$ differs too much from the current policy $\pi_{\theta^{(i-1)}}$: 
\begin{theorem}[Offline setting can fail with high probability]
\label{thm:1}
Let $\pi$ be any distribution for which there exists $\by \in Y$ such that $\pi(\by\mid x), \pi(\by\mid x,y)\le \epsilon$ for all $y \in (Y \setminus \by ) \cup \{y^*\}$ and $\pi_{\theta^{(i-1)}}(\by \mid x)\ge c$ for some $i \in \{1,2,\cdots,M\}$. Set $\pi^{(i)}=\pi$ for all $i \in \{1,2,\cdots,M\}$. Then, there exists  $\theta\in\Theta$ such that with probability at least $1 - 2\epsilon M$ (over the samples  of $\pi^{(i)}=\pi$), the following holds: $\pi^{}_{\theta}(y^*\mid x) \le 1-c$.
\end{theorem}
If the current policy and the sampling policy differ too much, it is possible that $\epsilon=0$ and $c\approx1.0$, for which Theorem~\ref{thm:1} can conclude $\pi_{\theta}(y^*\mid x)\approx0$ with probability $1$ for any number of steps $M$. When $\epsilon \neq 0$, the lower bound of the failure probability decreases towards zero as we increase $M$. Thus, it is important to make sure that $\epsilon\neq 0$ and $\epsilon$ is not too low. This is achieved by using the online setting, \textit{i.e.}, $\pi^{(i)}=\pi_{\theta^{(i)}}$.
Therefore, Theorem~\ref{thm:1} motivates us to use the online setting. Theorem~\ref{thm:2} confirms that we can indeed avoid this failure case in the online setting. 
\begin{theorem}[Online setting can avoid offline failure case]
    \label{thm:2}
   Let $\pi^{(i)}=\pi_{\theta^{(i-1)}}$. Then, for any $\theta\in \Theta$, it holds that $\pi_{\theta}(y^*\mid x) = 1$ if $M\ge n+1$.  
\end{theorem}
See Appendix~\ref{app:online-dpo} for the proofs of Theorems \ref{thm:1} and \ref{thm:2}. As suggested by the theorems, a better sampling policy is to use both the latest policy and the optimal policy for preference sampling. However, since we cannot access the optimal policy $\pi^*$ in practice, we adopt online DPO via sampling from the latest policy $\pi_{\theta^{(i-1)}}$. The key insight of our iterative preference learning approach is that online DPO is proven to enable us to converge to an optimal policy even if it is inaccessible to sample outputs. We provide further discussion and additional insights in Appendix~\ref{app:online-dpo}.

\section{Experiments}~\label{sec:exp}
We evaluate the effectiveness of MCTS-enhanced iterative preference learning on arithmetic and commonsense reasoning tasks. We employ Mistral-7B~\citep{DBLP:journals/corr/abs-2310-06825} as the base pre-trained model. We conduct supervised training using Arithmo \footnote{\href{https://huggingface.co/datasets/akjindal53244/Arithmo-Data}{https://huggingface.co/datasets/akjindal53244/Arithmo-Data}} which comprises approximately $540$K mathematical and coding problems. 
Detailed information regarding the task formats, specific implementation procedures, and parameter settings of our experiments can be found in Appendix~\ref{app:implementation}. 

\paragraph{Datasets.} We aim to demonstrate the effectiveness and versatility of our approach by focusing on two types of reasoning: arithmetic and commonsense reasoning. For arithmetic reasoning, we utilize two datasets: GSM8K~\citep{DBLP:journals/corr/abs-2110-14168}, which consists of grade school math word problems, and MATH~\citep{DBLP:conf/nips/HendrycksBKABTS21}, featuring challenging competition math problems. Specifically, in the GSM8K dataset, we assess both chain-of-thought (CoT) and program-of-thought (PoT) reasoning abilities. We integrate the training data from GSM8K and MATH to construct the prompt data for our preference learning framework, aligning with a subset of the Arithmo data used for Supervised Fine-Tuning (SFT). This approach allows us to evaluate whether our method enhances reasoning abilities on specific arithmetic tasks. For commonsense reasoning, we use four multiple-choice datasets: ARC (easy and challenge splits)~\citep{DBLP:journals/corr/abs-1803-05457}, focusing on science exams; AI2Science (elementary and middle splits)~\citep{DBLP:journals/corr/abs-1803-05457}, containing science questions from student assessments; OpenBookQA (OBQA)~\citep{DBLP:conf/emnlp/MihaylovCKS18}, which involves open book exams requiring broad common knowledge; and CommonSenseQA (CSQA)~\citep{talmor-etal-2019-commonsenseqa}, featuring commonsense questions necessitating prior world knowledge. The diversity of these datasets, with different splits representing various grade levels, enables a comprehensive assessment of our method's generalizability in learning various reasoning tasks through self-distillation. Performance evaluation is conducted using the corresponding validation sets of each dataset. Furthermore, we employ an unseen evaluation using the validation set of an additional dataset, SciQ~\citep{DBLP:conf/aclnut/WelblLG17}, following the approach of \citet{liu-etal-2023-crystal}, to test our model's ability to generalize to novel reasoning contexts. 

\textbf{Baselines.} Our study involves a comparative evaluation of our method against several prominent approaches and fair comparison against variants including instance-level iterative preference learning and offline MCTS-enhanced learning. We use instance-level sampling as a counterpart of step-level preference collection via MCTS. For a fair comparison, we also apply self-evaluation and correctness assessment and control the number of samples under a comparable compute budget with MCTS in instance-level sampling. The offline version uses the initial policy for sampling rather than the updated one at each iteration.

We contrast our approach with the Self-Taught Reasoner (STaR)\citep{DBLP:conf/nips/ZelikmanWMG22}, an iterated learning model based on instance-level rationale generation, and Crystal\citep{liu-etal-2023-crystal}, an RL-tuned model with a focus on knowledge introspection in commonsense reasoning. Considering the variation in base models used by these methods, we include comparisons with Direct Tuning, which entails fine-tuning base models directly bypassing chain-of-thought reasoning. In the context of arithmetic reasoning tasks, our analysis includes Language Model Self-Improvement (LMSI)\citep{DBLP:conf/emnlp/0001GHW00023}, a self-training method using self-consistency to gather positive data, and Math-Shepherd\citep{wang2023math}, which integrates process supervision within Proximal Policy Optimization (PPO). To account for differences in base models and experimental setups across these methods, we also present result performance of SFT models as baselines for each respective approach. 


\begin{table*}[t]
    \begin{minipage}[b]{.66\textwidth}
        \centering
\begin{center}
\tiny
\caption{Result comparison (accuracy $\%$) on arithmetic tasks. We supervised fine-tune the base model Mistral-7B on Arithmo data, while Math-Shepherd~\citep{wang2023math} use MetaMATH~\citep{yu2023metamath} for SFT. We highlight the advantages of our approach via conceptual comparison with other methods, where NR, OG, OF, and NS represent ``w/o Reward Model'', ``On-policy Generation'', ``Online Feedback'', and ``w/ Negative Samples''.
}
\label{tab:math}
\begin{tabular}{lcccccccc}
\toprule
\multirow{2}{*}{\textbf{Approach}} & \multirow{2}{*}{\textbf{Base Model}} & \multicolumn{4}{c}{\textbf{Conceptual Comparison}} & \multirow{2}{*}{\textbf{GSM8K}} & \multirow{2}{*}{\textbf{MATH}} \\
\cmidrule(lr){3-6}
& & NR & OG & OF & NS &  &  \\
\midrule
LMSI & PaLM-540B & \ding{51} & \ding{51} & \ding{55} & \ding{55} & $73.5$ & $-$ \\
\midrule
SFT (MetaMath) & \multirow{2}{*}{Mistral-7B} & $-$ & $-$ & $-$ & $-$ & $77.7$ & $28.2$ \\
Math-Shepherd & & \ding{55} & \ding{51} & \ding{55} & \ding{51} & $\mathbf{84.1}$ & $33.0$ \\
\midrule
SFT (Arithmo) & \multirow{4}{*}{Mistral-7B} & $-$ & $-$ & $-$ & $-$ & $75.9$ & $28.9$ \\
MCTS Offline-DPO & & \ding{51} & \ding{55} & \ding{55} & \ding{51} & $79.9$ & $31.9$ \\
Instance-level Online-DPO & & \ding{51} & \ding{51} & \ding{51} & \ding{51} & $79.7$ & $32.9$ \\
Ours & & \ding{51} & \ding{51} & \ding{51} & \ding{51} & $80.7$ & $32.2$ \\
Ours (w/ G.T.) & & \ding{51} & \ding{51} & \ding{51} & \ding{51} & $81.8$ & $\mathbf{34.7}$ \\
\bottomrule
\end{tabular}
\end{center}
    \end{minipage}
    \quad
    \begin{minipage}[b]{.31\textwidth}
        \centering
    \includegraphics[width=\textwidth]{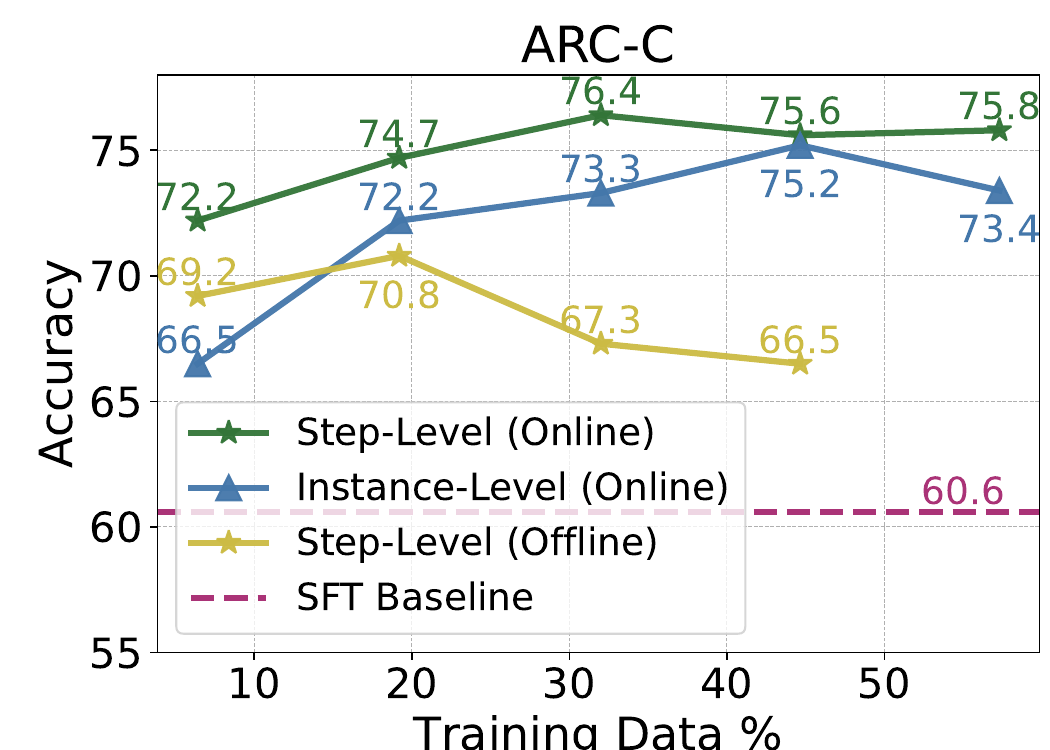}
    \vskip -0.1in
    \captionof{figure}{Performance on the validation set of ARC-C via training with different settings.}
    \label{fig:ablation-curves}
    \end{minipage}
    \vskip -0.15in
\end{table*}
\begin{table*}[t]
\centering
\tiny
\caption{Result comparisons (accuracy $\%$) on commonsense reasoning tasks. The results based on GPT-3-curie~\citep{brown2020language} and T5~\citep{raffel2020exploring} are reported from \citet{liu-etal-2023-crystal}. For CSQA, we also include the GPT-J~\citep{gpt-j} results reported by \citet{DBLP:conf/nips/ZelikmanWMG22}. We follow \citet{liu-etal-2023-crystal} to combine the training data of ARC, AI2Sci, OBQA, and CSQA for training
, while STaR~\citep{DBLP:conf/nips/ZelikmanWMG22} only use CSQA for training. }
\label{tab:sqa}
\begin{tabular}{lcccccccccc}
\hline
\multirow{2}{*}{\textbf{Approach}} & \multirow{2}{*}{\textbf{Base Model}} & \multicolumn{4}{c}{\textbf{Conceptual Comparison}} & \multirow{2}{*}{\textbf{ARC-c}} & \multirow{2}{*}{\textbf{AI2Sci-m}} & \multirow{2}{*}{\textbf{CSQA}} & \multirow{2}{*}{\textbf{SciQ}} & \multirow{3}{*}{\shortstack{\textbf{Train Data}\\\textbf{Used} ($\%$)}}  \\
\cmidrule(lr){3-6}
 & & NR & OG & OF & NS & & & & &  \\
\hline
CoT Tuning & GPT-3-curie (6.7B) & \ding{51} & \ding{55} & \ding{55} & \ding{55} & $-$ & $-$ & $56.8$ & $-$ & $100$ \\
\hline
Direct Tuning & \multirow{2}{*}{GPT-J (6B)} & \ding{51} & \ding{55} & \ding{55} & \ding{55} & $-$ & $-$ & $60.0$ & $-$ & $100$ \\
STaR &  & \ding{51} & \ding{51} & \ding{51} & \ding{55}& $-$ & $-$ & $72.5$ & $-$ & $86.7$ \\
\hline
Direct TUning & \multirow{2}{*}{T5-11B} & \ding{51} & \ding{55} & \ding{55} & \ding{55} & $72.9$ & $84.0$ & $82.0$ & $83.2$ & $100$ \\
Crystal &   & \ding{55} & \ding{51} & \ding{51} & \ding{51}& $73.2$ & $84.8$ & $\mathbf{82.3}$ & $85.3$ & $100$ \\
\hline
SFT Base (Arithmo) & \multirow{5}{*}{Mistral-7B} & $-$ & $-$ & $-$ & $-$ & $60.6$ & $70.9$ & $54.1$ & $80.8$ & $-$ \\
Direct Tuning &   & \ding{51} & \ding{55} & \ding{55} & \ding{55}& $73.9$ & $85.2$ & $79.3$ & $86.4$ & $100$ \\
MCTS Offline-DPO & & \ding{51} & \ding{55} & \ding{55} & \ding{51}& $70.8$ & $82.6$ & $68.5$ & $87.4$ & $19.2$ \\
Instance-level Online-DPO & & \ding{51} & \ding{51} & \ding{51} & \ding{51}& $75.3$ & $87.3$ & $63.1$ & $87.6$ & $45.6$ \\
Ours &  & \ding{51} & \ding{51} & \ding{51} & \ding{51}& $\mathbf{76.4}$ & $\mathbf{88.2}$ & $74.8$ & $\mathbf{88.5}$ & $47.8$ \\
\hline
\end{tabular}
\vskip -0.21in
\end{table*}

\subsection{Main Results}~\label{sec:result}
\textbf{Arithmetic Reasoning.}  In Table~\ref{tab:math}, we present a comparative analysis of performance gains in arithmetic reasoning tasks. Our method demonstrates substantial improvements, notably on GSM8K, increasing from $75.9\% \rightarrow 81.8\%$, and on MATH, enhancing from $28.9\% \rightarrow 34.7\%$. When compared to Math-Shepherd, which also utilizes process supervision in preference learning, our approach achieves similar performance enhancements without the necessity of training separate reward or value networks. This suggests the potential of integrating trained reward model signals into our MCTS stage to further augment performance. Furthermore, we observe significant performance gain on MATH when incorporating the ground-truth solutions in the MCTS process for preference data collection, illustrating an effective way to refine the preference data quality with G.T. guidance.


\begin{figure*}[t]
    \centering
    \begin{minipage}[b]{.32\linewidth}
        \centering
        \includegraphics[width=\textwidth]{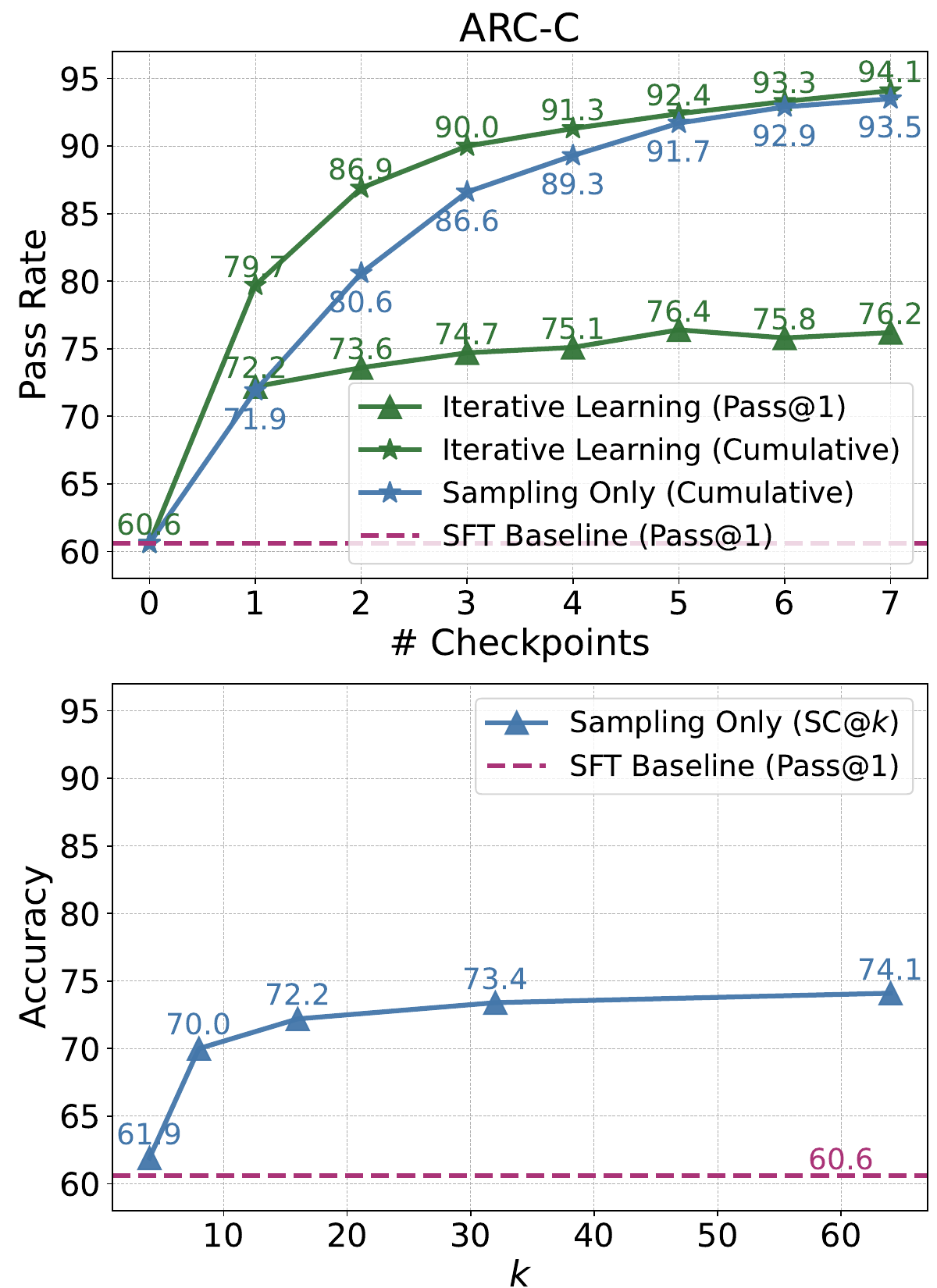}
    \end{minipage}
    \begin{minipage}[b]{.32\linewidth}
        \centering
         \includegraphics[width=\textwidth]{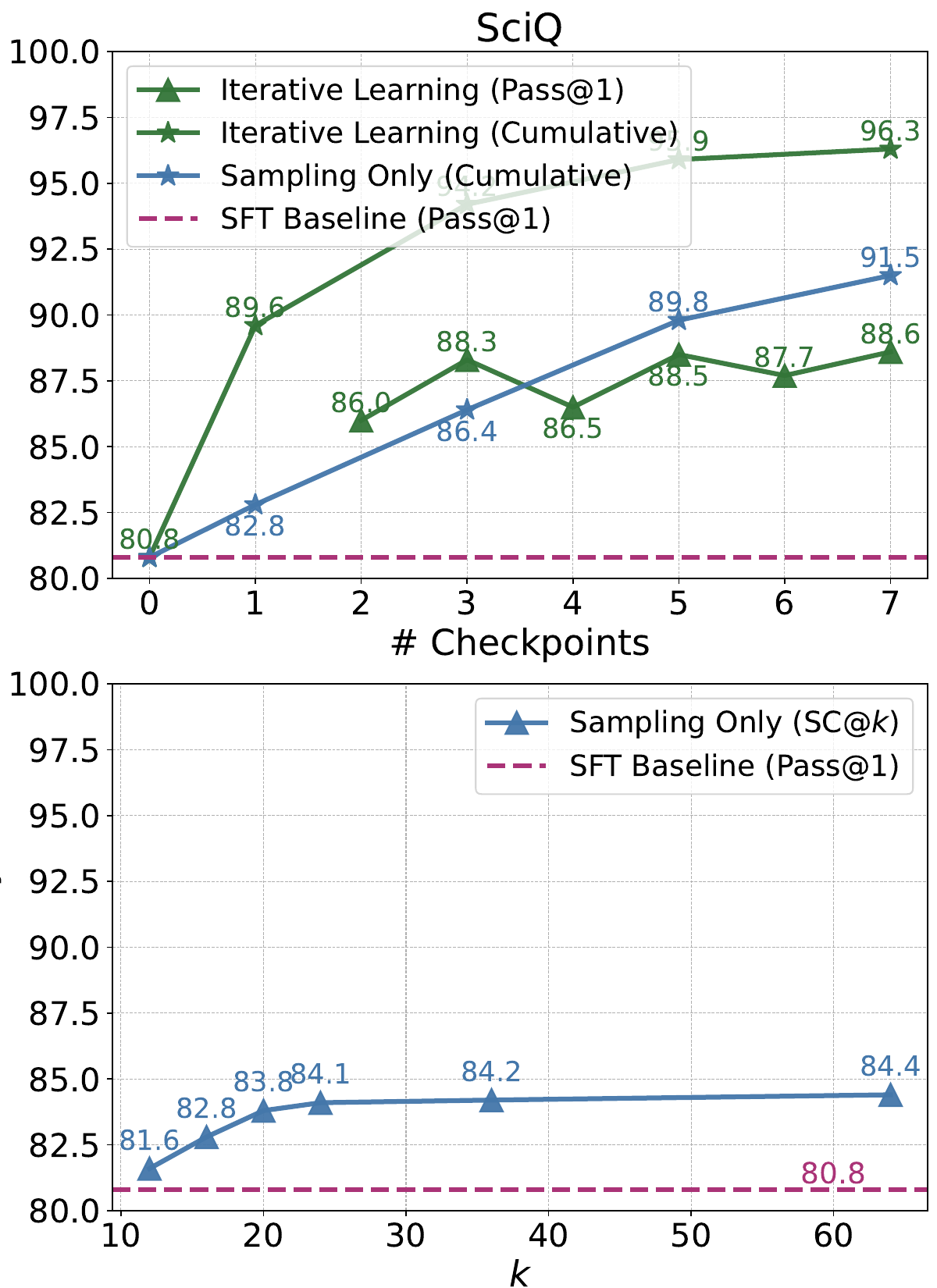}
    \end{minipage}
    \begin{minipage}[b]{.32\linewidth}
        \centering
        \includegraphics[width=\textwidth]{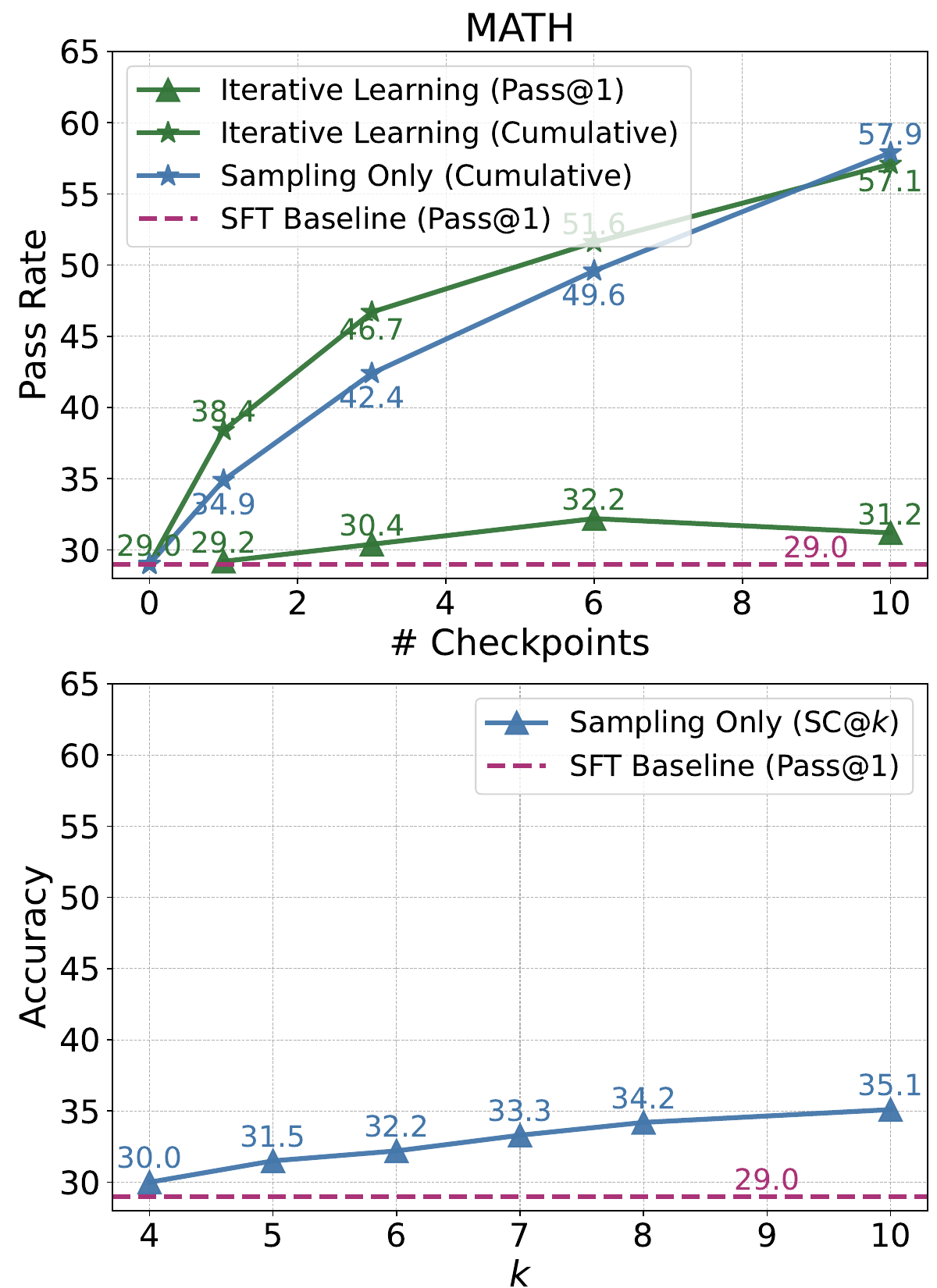}
    \end{minipage}
    \vskip -0.1in
    \caption{Training- vs. Test- Time Compute Scaling on ARC-C, SciQ, and MATH evaluation sets. The cumulative pass rate of our iterative learning method can be seen as the pass rate of an ensemble of different model checkpoints. We use greedy decoding to obtain the inference time performance of our method of iterative learning.}
    \label{fig:train-test}
    \vskip -0.1in
\end{figure*}

\begin{table*}[t]
\centering
\small
\caption{Ablation of ``EXAMPLE ANSWER'' in self-evaluation on GSM8K, MATH, and ARC-C. We report AUC and accuracy ($\%$) to compare the discriminative abilities of self-evaluation scores.}
\label{tab:exp-ans}
\begin{tabular}{lcccccc}
\toprule
\multirow{2}{*}{\textbf{Approach}} & \multicolumn{2}{c}{\textbf{GSM8K}} & \multicolumn{2}{c}{\textbf{MATH}} & \multicolumn{2}{c}{\textbf{ARC-C}} \\
\cmidrule(lr){2-3} \cmidrule(lr){4-5} \cmidrule(lr){6-7}
 & \textbf{AUC} & \textbf{Accuracy} & \textbf{AUC} & \textbf{Accuracy} & \textbf{AUC} & \textbf{Accuracy} \\
\midrule
w/ example answer & $74.7$ & $72.5$ & $76.6$ & $48.8$ & $65.2$ & $57.5$ \\
w/o example answer & $62.0$ & $69.5$ & $48.1$ & $42.3$ & $55.8$ & $48.4$ \\
\bottomrule
\end{tabular}
\vskip -0.2in
\end{table*}

\textbf{Commonsense Reasoning.} In Table~\ref{tab:sqa}, we report the performance on commonsense reasoning tasks, where our method shows consistent improvements. Notably, we achieve absolute accuracy increases of $2.5\%$, $3.0\%$, and $2.1\%$ on ARC-Challenge (ARC-C), AI2Sci-Middle (AI2Sci-M), and SciQ, respectively, surpassing the results of direct tuning. However, in tasks like OBQA and CSQA, our method, focusing on intermediate reasoning refinement, is less efficient compared to direct tuning. Despite significant improvements over the Supervised Fine-Tuning (SFT) baseline (for instance, from $59.8\%$ to $79.2\%$ on OBQA, and from $54.1\%$ to $74.8\%$ on CSQA), the gains are modest relative to direct tuning.
This discrepancy could be attributed to the base model's lack of specific knowledge, where eliciting intermediate reasoning chains may introduce increased uncertainty in model generations, leading to incorrect predictions. We delve deeper into this issue of hallucination and its implications in our qualitative analysis, as detailed in Section~\ref{sec:analysis}.


\subsection{Further Analysis}~\label{sec:analysis}
\textbf{Training- vs. Test- Time Compute Scaling.} Our method integrates MCTS with preference learning, aiming to enhance both preference quality and policy reasoning via step-level alignment. We analyze the impact of training-time compute scaling versus increased inference-time sampling.

We measure success by the pass rate, indicating the percentage of correctly elicited answers. Figure~\ref{fig:train-test} displays the cumulative pass rate at each checkpoint, aggregating the pass rates up to that point. For test-time scaling, we increase the number of sampled reasoning chains. Additionally, we compare the inference performance of our checkpoints with a sampling-only method, self-consistency, to assess their potential performance ceilings. The pass rate curves on ARC-C, SciQ, and MATH datasets reveal that our MCTS-enhanced approach yields a higher training compute scaling exponent. This effect is particularly pronounced on the unseen SciQ dataset, highlighting our method's efficiency and effectiveness in enhancing specific reasoning abilities with broad applicability.
Inference-time performance analysis shows higher performance upper bounds of our method on ARC-C and SciQ. For instance, while self-consistency on SciQ plateaus at around $84\%$, our framework pushes performance to $88.6\%$. However, on MATH, the sampling-only approach outperforms training compute scaling: more sampling consistently enhances performance beyond $35\%$, whereas post-training performance hovers around $32.2\%$. This observation suggests that in-domain SFT already aligns the model well with task-specific requirements. 

\paragraph{Functions of Self-Evaluation Mechanism.}
As illustrated in Section~\ref{sec:mcts}, the self-evaluation score inherently revises the $Q$ value estimation for subsequent preference data collection. In practice, we find that the ground-truth information, \textit{i.e.}, the ``EXAMPLE ANSWER'' in Table~\ref{tab:eval-prompt}, is crucial to ensure the reliability of self-evaluation. We now compare the score distribution and discriminative abilities when including v.s. excluding this ground-truth information in Table~\ref{tab:exp-ans}. With this information , the accuracy of self-evaluation significantly improves across GSM8K, MATH, and ARC-C datasets.

\paragraph{Ablation Study.}
We ablate the impact of step-level supervision signals and the online learning aspect of our MCTS-based approach. Tables~\ref{tab:math} and~\ref{tab:sqa} shows performance comparisons across commonsense and arithmetic reasoning tasks under different settings. Our method, focusing on step-level online preference learning, consistently outperforms both instance-level and offline approaches in commonsense reasoning. For example, we achieve $76.4\%$ on ARC-C and $88.5\%$ on SciQ, surpassing $70.8\%$ and $87.4\%$ of the offline variant, and $75.3\%$ and $87.6\%$ of the instance-level approach.


\begin{figure*}
    \centering
    \includegraphics[width=\linewidth]{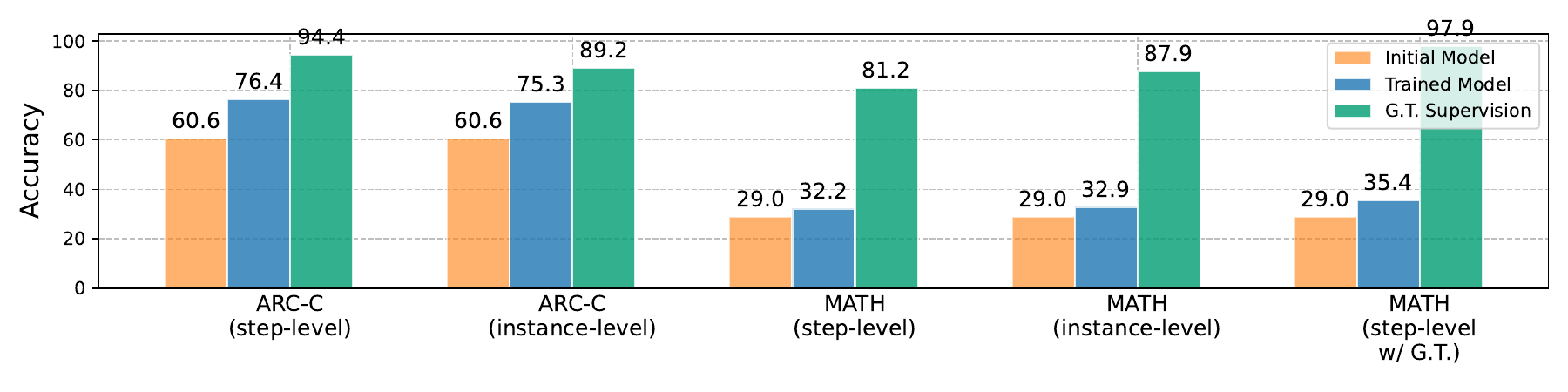}
    \vskip -0.2in
    \caption{Abltation study on step- vs. instance-level supervision, online vs. offline learning, and w/ vs. w/o G.T. supervision. We also compare the accuracy of the training signals collected via MCTS or instance-level sampling.}
    \label{fig:gt}
    \vskip -0.1in
\end{figure*}

\begin{table}[t]
\caption{Qualitative Analysis of model outputs on OBQA, AI2Sci, and MATH. We highlight \textcolor{ForestGreen}{correct} and \textcolor[RGB]{155,97,183}{incorrect} keywords. The predicted answers on MATH are \boxed{\textrm{boxed}}, where our method solves the complex problem while SFT baseline failed to elicit a complete reasoning chain.}
\label{tab:qualitative}
\vskip 0.15in
\begin{center}
\tiny
\begin{tabular}{llc}
\toprule
Task & Inputs and Outputs & Answer \\
\midrule
\multirow{8}{*}{OBQA} & \textbf{Question}: As the rain forest is deforested the atmosphere will increase with: & \multirow{8}{*}{(C)} \\
 & \textbf{Answer Choices}: (A) oxygen (B) nitrogen (C) carbon (D) rain & \\
 & \textbf{SFT Baseline}: As the rain forest is deforested the atmosphere will increase with carbon. The answer is \textcolor{ForestGreen}{C}. & \\
 & \textbf{Ours}: When trees are cut down, the \textcolor{ForestGreen}{carbon} stored in the trees is released into the atmosphere as carbon & \\
 &  dioxide. This carbon dioxide then increases the \textcolor{ForestGreen}{carbon} content of the atmosphere. {Since \textcolor{ForestGreen}{carbon} is the main}  & \\
 & {ingredient in plant life, the increased \textcolor{ForestGreen}{carbon} content of the atmosphere will lead to increased plant growth,}   & \\
 & {which will lead to increased photosynthesis, which will lead to increased \textcolor[RGB]{155,97,183}{oxygen} production.} Therefore,  & \\
 & \textcolor[RGB]{155,97,183}{oxygen} is the gas that will increase in the atmosphere as the rain forest is deforested. The answer is: \textcolor[RGB]{155,97,183}{A}. & \\
 \midrule
 \multirow{9}{*}{AI2Sci} & \textbf{Question}: Which pair of instruments can be used to measure wind chill? & \multirow{9}{*}{(D)} \\
  & \textbf{Answer Choices}: (A) barometer and wind vane (B) wind vane and anemometer  & \\
  & (C) thermometer and barometer (D) anemometer and thermometer & \\
  & \textbf{SFT Baseline}: Wind chill is the effect of wind on the apparent temperature, it is the temperature that the human  & \\
  & body feels due to the movement of air. It is an index of the discomfort felt by the human body due to low  & \\
  & temperature and high wind speed. It is measured by an instrument called \textcolor[RGB]{155,97,183}{wind chill meter}. The answer is \textcolor[RGB]{155,97,183}{B} & \\
  & \textbf{Ours}: To measure wind chill, you would need an \textcolor{ForestGreen}{anemometer and a thermometer}. The anemometer measures  & \\
  & the wind speed, and the thermometer measures the temperature. Together, these two instruments can be used to  & \\
  & calculate the wind chill factor. The answer is: \textcolor{ForestGreen}{D} & \\
\midrule
\multirow{6}{*}{MATH} & \textbf{Question}: Let $f(x) = x^7 + ax^3 + cx$  If $f(5) = 3,$ then find $f(-5)$.  & \multirow{6}{*}{-3}  \\
 & \textbf{SFT Baseline}: We know that $f(5) = 5^7 + 5a + 5c = 3.$ We want to find $f(-5),$ which is $(-5)^7 + (-5a)$  & \\
 & $+ (-5c).$ Since $(-5)^7 = -5^7 = -3,$ we have $f(-5) = -3 - 5a - 5c.$ We can rewrite this as & \\
 & $f(-5) = -3 - 5(-3) - 5c = -3 + 15 - 5c = \boxed{\textcolor[RGB]{155,97,183}{12-5c}}.$ & \\
 & \textbf{Ours}: We know that $f(5) = 5^7 + 5a + 5c = 3.$ Since $f(x)$ is a polynomial, we can use the fact that  & \\
 & $f(-5) = (-5)^7 - 5a - 5c = \boxed{\textcolor{ForestGreen}{-3}}.$ & \\
\midrule
\multirow{14}{*}{MATH} & \textbf{Question}: If $f(c)=\frac{3}{2c-3}$, find $\frac{kn^2}{lm}$ when $f^{-1}(c)\times c \times f(c)$ equals the simplified fraction$\frac{kc+l}{mc+n}$,  & \multirow{14}{*}{9}  \\
 & where $k,l,m,\text{ and }n$ are integers. & \\
 & \textbf{SFT Baseline}: We start by finding $f^{-1}(c)$. To do this, we set $f(c) = x$ and solve for $c$: $\frac{3}{2c-3} = x $,  & \\
 & $2c-3 = \frac{3}{x}$, $2c = \frac{3}{x} + 3$, $c = \frac{\frac{3}{x} + 3}{2}$.So $f^{-1}(c) = \frac{\frac{3}{c} + 3}{2c}$. Next, we substitute $f^{-1}(c)$ into the expression  & \\
 & $f^{-1}(c) \times c \times f(c)$: $\frac{\frac{3}{c} + 3}{2c} \times c \times \frac{3}{2c-3}$ Simplifying, we get: $\frac{3 + 3c}{2c} \times \frac{3}{2c-3}$ $\frac{9 + 9c}{4c^2 - 6c}$ $\frac{9 + 9c}{4c(c-1)}$  & \\
 & We want this expression to be equal to $\frac{kc+l}{mc+n}$, so we set them equal to each other and solve for $k,l,m,$  & \\
 & and $n$: $\frac{9 + 9c}{4c(c-1)} = \frac{kc+l}{mc+n}$ Cross-multiplying, we get: $9 + 9c = 4c(c-1)(kc+l)$ Expanding and simplifying,  & \\
 & we get: $9 + 9c = 4c^2k - 4ck - 4cl + l$ Rearranging terms, we get: $4c^2k - 4$ & \\
 & \textbf{Ours}: We know that $f(f^{-1}(c))=c$ for all $c$ in the domain of $f^{-1}$. So, $f(f^{-1}(c))=f(c)$ for all $c$ in the  & \\
 & domain of $f^{-1}$. Therefore, $f(f^{-1}(c))=f(c)$ implies $f(c)=c$ for all $c$ in the domain of $f^{-1}$. Since  & \\
 & $f(c)=\frac{3}{2c-3}$, we have $f(c)=c$ for all $c$ in the domain of $f^{-1}$. This means that $2c-3=c$ for all $c$ in the  & \\
 & domain of $f^{-1}$. Simplifying, we get $c=3$ for all $c$ in the domain of $f^{-1}$. Therefore, $f^{-1}(c)=3$ for all $c$  & \\
 & in the domain of $f^{-1}$. Now, we can substitute $f^{-1}(c)=3$ into $f^{-1}(c)\times c \times f(c)$ to get $3\times 3 \times \frac{3}{2(3)-3}$.  & \\
 & Simplifying, we get $9 \times \frac{3}{3}=9$. Therefore, $\frac{kn^2}{lm}=\boxed{\textcolor{ForestGreen}{9}}$ & \\ 
\bottomrule
\end{tabular}
\end{center}
\vskip -0.2in
\end{table}

In arithmetic reasoning, performance differences among settings are less pronounced for challenging task such as MATH without the incorporation of ground-truth solutions (\textit{e.g.}, $32.2\%$ for our method (w/o G.T.) vs. $31.9\%$ and $32.9\%$ for offline and instance-level on MATH). The comparable performance of offline learning aligns with our theoretical analysis that offline approaches can be effective when the initial policy is already well-tuned with high-quality, in-domain data. We further interpret how G.T. guidance integration to enhance the effectiveness of our framework in Figure~\ref{fig:gt}. With G.T. supervision, the accuracy of training signals improve significantly from $81.2\%$ to $97.9\%$, leading to substantial performance gain on model performance. This also explains the similar performance (w/o G.T.) between corresponding using step- and instance-level supervision, where our step-level approach shows effectiveness in narrowing the gap between accuracies of corresponding supervisions.

\paragraph{Training Dynamics in Iterative Learning.}
As shown in Figure~\ref{fig:ablation-curves}, online learning exhibits cyclic performance fluctuations, with validation performance peaking before dipping. We conduct theoretical analysis on this in Appendix~\ref{app:online-dpo} and shows that continuous policy updates with the latest models can lead to periodic knowledge loss due to insufficient optimization in iterative updates. We further probe these phenomena qualitatively next.

\paragraph{Qualitative Analysis.}
Our qualitative analysis in Table~\ref{tab:qualitative} examines the impact of step-level supervision on intermediate reasoning correctness across different tasks. In OBQA, the implementation of MCTS, as discussed in Section~\ref{sec:result}, often leads to longer reasoning chains. This can introduce errors in commonsense reasoning tasks, as seen in our OBQA example, where an extended chain results in an incorrect final prediction. Conversely, in the MATH dataset, our approach successfully guides the model to rectify mistakes and formulates accurate, extended reasoning chains, demonstrating its effectiveness in complex math word problems. This analysis underscores the need to balance reasoning chain length and logical coherence, particularly in tasks with higher uncertainty, such as commonsense reasoning.

\section{Related Work}~\label{sec:related}
Various studies focus on self-improvement to exploit the model's capability. 
One line of work focuses on collecting high-quality positive data from model generations guided by static reward heuristic~\citep{DBLP:conf/nips/ZelikmanWMG22, DBLP:journals/corr/abs-2308-08998, DBLP:conf/iclr/PoluHZBBS23}. 
Recently, \citet{yuan2024self} utilize the continuously updated LLM self-rewarding to collect both positive and negative data for preference learning. \citet{DBLP:journals/corr/abs-2306-13394} adopt exploration strategy via rejection sampling to do online data collection for iterative preference learning. Different from prior works at instance-level alignment, we leverage MCTS as a policy improvement operator to iteratively facilitate step-level preference learning. We discuss additional related work in Appendix~\ref{app:related}.

\section{Conclusion}~\label{sec:conclusion}
In this paper, we propose MCTS-enhanced iterative preference learning, utilizing MCTS as a policy improvement operator to enhance LLM alignment via step-level preference learning. MCTS balances quality exploitation and diversity exploration to produce high-quality training data, efficiently pushing the ceiling performance of the LLM on various reasoning tasks. Theoretical analysis shows that online sampling in our iterative learning framework is key to improving the LLM policy toward optimal alignment. We hope our proposed approach can inspire future research on LLM alignment from both data-centric and algorithm-improving aspects: to explore searching strategies and utilization of history data and policies to augment and diversify training examples; to strategically employ a tradeoff between offline and online learning to address the problem of cyclic performance change of the online learning framework as discussed in our theoretical analysis.

\begin{ack}
The computational work for this article was partially performed on resources of the National Supercomputing Centre (NSCC), Singapore\footnote{\href{https://www.nscc.sg/}{https://www.nscc.sg/}}. 
\end{ack}


\small
\bibliography{example_paper}
\bibliographystyle{acl_natbib}

\newpage
\appendix
\appendix
\onecolumn

\section{Related Work}~\label{app:related}
\paragraph{Iterated Learning.}
Typical iterated learning operates in a multi-agent scenario, consisting of a loop where an apprentice self-plays, learns from expert feedback, and replaces the current expert for the new iteration~\citep{DBLP:conf/nips/AnthonyTB17}. 
\citet{DBLP:conf/iclr/PoluHZBBS23} apply expert iteration on formal mathematical reasoning to conduct proof search interleaved with learning.
\citet{DBLP:conf/nips/ZelikmanWMG22} avoid the need for training a separate value function by directly assessing the final outcomes of reasoning to filter generated examples for iterated learning.
Recently, \citet{yuan2024self} leverage the technique of LLM-as-a-Judge~\citep{DBLP:journals/corr/abs-2306-05685} and introduce self-rewarding language models to improve LLM alignment with self-feedback. Differently, we combine the feedback of outcome assessment and LLM self-evaluation and further decompose them into fine-grained signals via MCTS for step-level iterative preference learning. 

\paragraph{Self-Training.}
Self-training uses unlabeled data to improve model training by assigning pseudo labels from a learned labeler~\citep{DBLP:journals/tit/Scudder65a, DBLP:conf/acl/Yarowsky95, DBLP:conf/cvpr/XieLHL20, DBLP:conf/iclr/HeGSR20, park2020improved, zoph2020rethinking}. 
Recent research has explored several alternatives to label the examples. 
\citet{DBLP:conf/nips/ZelikmanWMG22} and \citet{DBLP:journals/corr/abs-2308-08998} use static reward heuristic to curate high-quality examples, while \citet{DBLP:conf/emnlp/0001GHW00023} collect high-confidence outputs as training data via chain-of-thought prompting~\citep{DBLP:conf/nips/Wei0SBIXCLZ22} and self-consistency~\citep{DBLP:conf/iclr/0002WSLCNCZ23}.
\citet{DBLP:journals/corr/abs-2309-00267} and \citet{yuan2024self} utilize the off-the-shelf LLM to reward its generations for preference learning.
To mitigate the noise from the sparse instance-level signals, we further refine the preference labels via stepwise tree search and LLM self-evaluation.

\paragraph{Preference Learning.}
The empirical achievements of LLMs have identified the benefits of RL techniques to better align with human preferences~\citep{DBLP:journals/corr/abs-2307-09288, DBLP:journals/corr/abs-2009-01325, DBLP:conf/nips/Ouyang0JAWMZASR22, DBLP:journals/corr/abs-2204-05862}.
The standard preference learning process learns a reward model to provide feedback in online RL~\citep{DBLP:journals/corr/SchulmanWDRK17}.
Recently, a variety of studies avoid training separate reward or value networks by hindsight instruction relabeling~\citep{DBLP:conf/icml/ZhangLWAG23}, direct preference optimization~\citep{DBLP:journals/corr/abs-2305-18290} and LLM self-evaluation~\citep{ren2023self}.
We further explore automatic supervision with MCTS to collect step-level preferences by breaking down outcome correctness integrated with self-evaluation. 
Our approach enables the continual collection of better-quality data via iterative learning, mitigating the limit of preference data when using a frozen reward model or offline learning algorithms.

\paragraph{Guided Search for Reasoning.}
Recent works improve the LLM reasoning ability by eliciting the intermediate reasoning chain~\citep{DBLP:conf/nips/Wei0SBIXCLZ22} and breaking it down into multiple steps via searching~\citep{DBLP:journals/corr/abs-2305-10601, DBLP:conf/emnlp/HaoGMHWWH23, DBLP:journals/corr/abs-2311-09724}.
The decomposition of the reasoning process has also been shown effective in reinforcement learning.
\citet{DBLP:journals/corr/abs-2305-20050} and \citet{li2023making} apply process supervision to train more reliable reward models than outcome supervision in mathematical reasoning~\citep{DBLP:journals/corr/abs-2211-14275}.
\citet{wang2023math} reinforce LLMs step-by-step with process supervision data automatically collected via model sampling and annotation.
We leverage the look-ahead ability of MCTS and integrate it with step-by-step self-evaluation to provide refined process supervision for reasoning.
This improves the generalization ability of our framework to update the policy via real-time collected preferences iteratively.

\section{Theoretical Analysis of Online DPO}~\label{app:online-dpo}
\paragraph{Preliminaries.}
A typical alignment technique begins with a policy $\pi_{\mathrm{sft}}(y\mid x)$ supervisedly fine-tuned on high-quality data from the target domain, where $x$ and $y$ are the prompt and the response, respectively.
The SFT policy is used to sample pairs of responses $(y_1, y_2) \sim \pi_{\mathrm{sft}}(y\mid x)$ with prompts $x$, which will be further labeled as pairwise preference data $y_w \succ y_l \mid x$, where $y_w$ and $y_l$ represent the preferred and dispreferred responses respectively.
The standard RLHF paradigm trains a reward model~\citep{DBLP:conf/nips/Ouyang0JAWMZASR22} on the preference data and employs PPO~\citep{DBLP:journals/corr/SchulmanWDRK17} to optimize the policy $\pi_{\theta}$ with the feedback provided by the reward model, where $\pi_{\theta}$ is also initialized to $\pi_{\mathrm{sft}}$ in practice.
DPO avoids fitting a reward model by optimizing the policy $\pi_{\theta}$ using preferences directly.

Given a reward function $r(x, y)$ and prompt distribution $\mathcal{P}$, RLHF and DPO optimize the KL-constrained reward maximization objective as follows:
\begin{equation}
    \label{eq:objective}
    \max_{\pi}\mathbb{E}_{x\sim \mathcal{P}, y\sim \pi}[r(x,y)] - \beta\mathbb{D}_{\mathrm{KL}}[\pi(y\mid x)\parallel \pi_{\mathrm{sft}}(y\mid x)]
\end{equation}
where $\beta$ scales the strength of the KL constraint. 
Let the ground-truth reward function be $r^*$, then \citet{DBLP:journals/corr/abs-2305-18290} estimate the optimal policy $\pi^*$ by fitting the Bradley-Terry model~\citep{bradley1952rank} on preference data:
\begin{equation}
    \label{eq:preference}
    \begin{aligned}
    & p^*(y_1\succ y_1\mid x) = \sigma(r^*(x,y_1)-r^*(x,y_2)) \\
    = & \frac{1}{1 + \exp\left(\beta\log\frac{\pi^*(y_2\mid x)}{\pi_{\mathrm{sft}}(y_2\mid x)}-\beta\log\frac{\pi^*(y_1\mid x)}{\pi_{\mathrm{sft}}(y_1\mid x)}\right)}
    \end{aligned}
\end{equation}
As the maximum likelihood estimator (MLE) of the optimal policy requires preferences sampled from the target policy~\citep{DBLP:journals/corr/abs-2309-06657}, DPO uses a fixed, potentially optimal but unknown policy to collect preference data of good quality.
This discrepancy can be a problem when the sampling policy differs dramatically from the current policy.
Moreover, the absence of a reward model in DPO presents challenges in learning from additional policy-generated data that lacks explicit preference indicators.
We further discuss the offline and online settings of DPO in Section~\ref{sec:theory}.

\paragraph{Additional details on labeling outcomes.} 
After a pair of outcomes $(y^{(i)},{y^{\prime}}^{(i)})$ are sampled from some sampling policy $\pi^{(i)}$, these are labeled to be $y_{w}^{(i)}$ and $y_l^{(i)}$ according to some preference density $p$. That is,  $\Pr[(y_w ^{(i)},y_{l}^{(i)})= (y^{(i)},{y^{\prime}}^{(i)})]=p(y^{(i)}\succ y'^{(i)} \mid x)$ and $\Pr[(y_w^{(i)} ,y_{l}^{(i)})= ({y^{\prime}}^{(i)},y^{(i)})]=1-p(y^{(i)}\succ {y^{\prime}}^{(i)} \mid x)$. For simplicity, a preference density is set to be $p(y^* \succ \by \mid x)=1$ for every optima-suboptimal pairs $(y^*, \by)$ for all $\by \in Y$.
We do not specify the  preference density for other pairs, \textit{i.e.}, $p(\by \succ \by^{\prime} \mid x)$ is arbitrary 
for $(\by, \by^{\prime}) \in Y \times Y$.

\paragraph{Abstract formulation for both offline and online settings.}
Our abstract formulation covers both the offline and online settings. The offline setting in previous papers is obtained by setting $\pi^{(i)} $ to be a single distribution fixed over $i\in \{1,2,\cdots,M\}$, \textit{e.g.}, an initial policy, an optimal policy, or an empirical data distribution of a given preference data. In the case of the empirical data distribution, the preference density  $p$ is set to the function outputting only $0$ or $1$ to recover the given preference data. 
The online setting is obtained by setting  $\pi^{(i)}=\pi_{\theta^{(i-1)}}$ where $\pi_{\theta^{(i-1)}}$ is the latest policy at the beginning of the $i$-th iteration, i.e., for $i\ge1$, ${\theta^{(i)}}^{}$ satisfies $\ell_j(\theta^{(i)})= 0$ for $j\in \{1, 2,\cdots,i-1\}$ and $\theta^{(0)}$ is the initialization. 
Thus, we can analyze both offline and online settings with this abstract framework.

\paragraph{Proof of Theorem \ref{thm:1}.}
\begin{proof}
The intuition behind the proof of Theorem \ref{thm:1} is that the current policy   $\pi_{\theta^{(i)}}$ may not be corrected if a fixed sampling policy $\pi$ never samples a suboptimal output $\by \in Y$ whose probability is high for the current policy  $\pi_{\theta^{(i)}}$. Let $\by$ be the suboptimal output such that $\pi(\by\mid x) \le \epsilon$ and $\pi_{\theta^{(i)}}(\by \mid x)\ge c$ for some $i \in \{1, 2, \cdots, M\}$. 
Denote preferences sampled by policy $\pi^{(i)}$ as $(y_w^{(i)}, y_l^{(i)})$.
From the definition of  the logistic function, we can rewrite
\begin{align*}
\ell_{i}(\theta) &=-\log \sigma \left(\beta \log  \frac{\pi_{\theta}(y_w^{(i)}\mid x)}{\pir(y_w^{(i)}\mid x)} -\beta \log  \frac{\pi_{\theta}(y_l^{(i)}\mid x)}{\pir(y_l^{(i)}\mid x)}\right)
\\ &=-\log \frac{1}{1+\exp(\beta \log  \frac{\pi_{\theta}(y_l^{(i)}\mid x)}{\pir(y_l^{(i)}\mid x)}-\beta \log  \frac{\pi_{\theta}(y_w^{(i)}\mid x)}{\pir(y_w^{(i)}\mid x)} )}  
\\ &=-\log \frac{\exp(\beta \log  \frac{\pi_{\theta}(y_w^{(i)}\mid x)}{\pir(y_w^{(i)}\mid x)})}{\exp(\beta \log  \frac{\pi_{\theta}(y_w^{(i)}\mid x)}{\pir(y_w^{(i)}\mid x)})+\exp(\beta \log  \frac{\pi_{\theta}(y_l^{(i)}\mid x)}{\pir(y_l^{(i)}\mid x)} )}
\\ &=-\log \frac{   \frac{\pi_{\theta}(y_w^{(i)}\mid x)^{\beta}}{\pir(y_w^{(i)}\mid x)^{\beta}}}{\frac{\pi_{\theta}(y_w^{(i)}\mid x)^{\beta}}{\pir(y_w^{(i)}\mid x)^{\beta}}+   \frac{\pi_{\theta}(y_l^{(i)}\mid x)^{\beta}}{\pir(y_l^{(i)}\mid x)^{^{\beta}}} }
\\ &=-\log \frac{ \pi_{\theta}(y_w^{(i)}\mid x)^{\beta}}{\pi_{\theta}(y_w^{(i)}\mid x)^{\beta}+ \pi_{\theta}(y_l^{(i)}\mid x)^{\beta}  (\frac{\pir(y_w^{(i)}\mid x)}{\pir(y_l^{(i)}\mid x)})^{\beta}}.
\end{align*}
From this equation, we observe that $\ell_{i}(\theta)$ can be minimized to be zero by minimizing $\pi_{\theta}(y_l^{(i)}\mid x)$ to be zero without maximizing $\pi_{\theta}(y_w^{(i)}\mid x)$. That is, for any $\beta>0$, if $\pi_{\theta}(y_l^{(i)}\mid x)=0$,
$$
\ell_{i}(\theta)=-\log \frac{ \pi_{\theta}(y_w^{(i)}\mid x)^{\beta}}{\pi_{\theta}(y_w^{(i)}\mid x)^{\beta}+0}=-\log 1=0.
$$ 
Thus, even if we sample $y^*$ with the optimal policy, $\ell_{i}(\theta)$ can be minimized without maximizing $\pi^{}_{\theta}(y^*\mid x)$ and minimizing $\pi^{}_{\theta}(\by|x)$ for $\by \neq y_l^{(i)}$. Thus, if $\by \neq y_l^{(i)}$ for all $i \in \{1, 2, \cdots, M\}$,  there exists $\theta$ such that $\ell_{i}(\theta) \le 0$ for all $i=1,\dots,M$, and 
$$
\pi_{\theta}^{}(\by \mid x)\ge c,
$$
because of the condition that  $\pi_{\theta}(\by \mid x)\ge c$ for some $i \in \{1, 2, \cdots, M\}$: i.e., $\pi_{\theta}(\by \mid x)$ is never minimized from the $i$-th iteration while minimizing $\ell_{i}(\theta)$ arbitrarily well, if  $\by$ is never sampled. 

Therefore, if  $\by$ is never sampled over $m$ iterations, since the probabilities sums up to one, we have 
$$
\pi^{}_{\theta}(y^*\mid x) \le1 - \pi_{\theta}^{}(\by |x)\le1 -c .
$$     
Moreover, 
$$ 
\Pr[\text{\ $\by$ being never sampled over $m$ iterations }] \ge (1-2\epsilon)^m \ge 1-2\epsilon m, 
$$
where the last line follows Bernoulli's inequality. By combining the above two equations, it holds that 
$$
\Pr[\pi^{}_{\theta}(y^*|x) \le 1-c] \ge  1 -2 \epsilon M.
$$ 
\end{proof}

\paragraph{Proof of Theorem \ref{thm:2}.}
\begin{proof}  From the proof of Theorem \ref{thm:1}, we have 
$$
\ell_{i}(\theta) =-\log \frac{ \pi_{\theta}(y_w^{(i)}\mid x)^{\beta}}{\pi_{\theta}(y_w^{(i)}\mid x)^{\beta}+ \pi_{\theta}(y_l^{(i)}\mid x)^{\beta}  (\frac{\pir(y_w^{(i)}\mid x)}{\pir(y_l^{(i)}\mid x)})^{\beta}}.
$$
For $\alpha\ge0$ and $\beta>0$, the condition $\ell_{i}(\theta) \le \alpha$ implies that 
\begin{align*}
&-\log \frac{ \pi_{\theta}(y_w^{(i)}\mid x)^{\beta}}{\pi_{\theta}(y_w^{(i)}\mid x)^{\beta}+ \pi_{\theta}(y_l^{(i)}\mid x)^{\beta}  (\frac{\pir(y_w^{(i)}\mid x)}{\pir(y_l^{(i)}\mid x)})^{\beta}} \le \alpha
\\& \Longleftrightarrow  \frac{ \pi_{\theta}(y_w^{(i)}\mid x)^{\beta}}{\pi_{\theta}(y_w^{(i)}\mid x)^{\beta}+ \pi_{\theta}(y_l^{(i)}\mid x)^{\beta}  (\frac{\pir(y_w^{(i)}|x)}{\pir(y_l^{(i)}\mid x)})^{\beta}} \ge  \exp(-\alpha)
\\& \Longleftrightarrow\pi_{\theta}(y_w^{(i)}\mid x)^{\beta} \ge  \exp(-\alpha)\pi_{\theta}(y_w^{(i)}\mid x)^{\beta}+\exp(-\alpha)\pi_{\theta}(y_l^{(i)}\mid x)^{\beta}  \left(\frac{\pir(y_w^{(i)}\mid x)}{\pir(y_l^{(i)}|x)}\right)^{\beta}
\\& \Longleftrightarrow\pi_{\theta}(y_w^{(i)}\mid x)^{\beta} [1-\exp(-\alpha)]\ge  \pi_{\theta}(y_l^{(i)}\mid x)^{\beta}  \exp(-\alpha)\left(\frac{\pir(y_w^{(i)}\mid x)}{\pir(y_l^{(i)}\mid x)}\right)^{\beta} \\& \Longleftrightarrow\pi_{\theta}(y_w^{(i)}\mid x)^{\beta} [1-\exp(-\alpha)]\ge  \pi_{\theta}(y_l^{(i)}\mid x)^{\beta}  \exp(-\alpha)\left(\frac{\pir(y_w^{(i)}\mid x)}{\pir(y_l^{(i)}\mid x)}\right)^{\beta} \\& \Longleftrightarrow\pi_{\theta}(y_w^{(i)}\mid x) (\exp(\alpha)-1)^{1/\beta}\left(\frac{\pir(y_l^{(i)}\mid x)}{\pir(y_w^{(i)}\mid x)}\right)^{}\ge  \pi_{\theta}(y_l^{(i)}\mid x).^{}   
\end{align*}
Since $\pi_{\theta}(y_w^{(i)}\mid x) \le 1$, this  implies that
\begin{align*}
\pi_{\theta}(y_l^{(i)}\mid x) &\le\pi_{\theta}(y_w^{(i)}\mid x) (\exp(\alpha)-1)^{1/\beta}\frac{\pir(y_l^{(i)}\mid x)}{\pir(y_w^{(i)}\mid x)}^{}
\\ & \le(\exp(\alpha)-1)^{1/\beta}\frac{\pir(y_l^{(i)}\mid x)}{\pir(y_w^{(i)}\mid x)}.
\end{align*} 

Thus, while we can prove a similar statement for $\alpha>0$ with this equation,  we set $\alpha=0$ for this theorem for a cleaner insight, yielding the following: the condition   $\ell_{i}(\theta) \le 0$  implies that
$$
\pi_{\theta}(y_l^{(i)}\mid x) =0.
$$
Since $y^{(i)}$ and ${y^{\prime}}^{(i)}$ are sampled from $\pi_{\theta^{(i)}}$ without replacement, this means that we  have $\pi_{\theta^{(i+k)}}(y_l^{(i)}\mid x) =0$ for all $k\ge1$ from the definition of  $\pi_{\theta^{(i)}}$: i.e.,  $\pi_{\theta^{(i)}}$ is the policy such that $\ell_j(\theta^{(i)})= 0$ for all $j=1,\dots,i-1$. Since $\pi_{\theta^{(i+k)}}^{}$ is then used to sample  $y^{(i)}$ and ${y^{\prime}}^{(i)}$ in the followings iterations for $k \ge 1$, we will never sample this $y_l^{(i)}$ again. Thus, at each iteration, we always sample pairs of  $y$ and $y^{\prime}$ such that these do not include an output judged to be not preferred in a previous iteration. 
This implies that at each iteration, we increase the number of suboptimal samples  $\by \in Y$ such that   $\pi_{\theta^{(i)}}(\by\mid x) =0$. In other words, we have
$$
|\{\by \in Y \mid \pi_{\theta^{(i)}}(\by\mid x)=0\}\mid \ge i-1.
$$   
Thus,        
$$
\pi_{\theta^{(i)}}(y^{*}\mid x) = 1- \sum_{j=1}^n \pi_{\theta^{(i)}}(\by_j\mid x)=1- \sum_{j \in S} \pi_{\theta^{(i)}}(\by_j\mid x). 
$$
where $|S|\le n+1-i$. Therefore, $\pi_{\theta^{(i)}}(y^{*}\mid x)=1$ when $i\ge n+1$.
 
\end{proof}

\paragraph{Additional discussion.}
We list the additional insights gained from the theoritical analysis.

$\bullet$ The proofs of Theorems \ref{thm:1}--\ref{thm:2} suggest that a better sampling policy is to use both the current policy and the optimal policy at the same time in the preference learning loss, i.e., sample $y \sim \pi^*$ and $y' \sim \pi_{\theta^{(i-1)}}$.\ This avoids the failure case of Theorem \ref{thm:1} and improves the convergence speed in Theorem \ref{thm:2}. However, since we cannot access the optimal policy $\pi^*$  in practice, Theorems \ref{thm:1}--\ref{thm:2} motivate online DPO. Online DPO is proven to enable us to converge to an optimal policy even if we cannot sample outputs from the optimal policy. 

$\bullet$ The proofs of Theorems \ref{thm:1}--\ref{thm:2} suggest that if we can sample from the optimal policy, then we can also use the samples of the optimal policy with the negative log-likelihood loss $-\log \pi_\theta(y^*\mid x)$ instead of DPO loss to avoid the failure case. 

$\bullet$ The proofs of Theorems \ref{thm:1}--\ref{thm:2} suggest that in the online setting, we should minimize the DPO loss to a certain low degree per iteration, i.e., we should take several rounds of minimization of DPO loss per online iteration, instead of only taking one round of minimization per iteration. This is because the proofs of Theorems \ref{thm:1}--\ref{thm:2} show that we can get into the cyclic situation in the online setting if the DPO loss is not minimized sufficiently per iteration. For example, we can sample $\by_1$ and $\by_2$ in one iteration and $\by_2$ and $\by_3$ in another iteration where  $\by_1 \succ\by_2 \succ \by_3$. If the probability of sampling $\by_{2}$ is not minimized sufficiently in the first iteration, it can be sampled again in the second iteration, where the probability of sampling $\by_{2}$ can be increased as $\by_2 \succ \by_3$. Then, this can repeat indefinitely. Thus, it is important to minimize DPO loss with several optimizer iterations per iteration.

\section{Implementation Details}~\label{app:implementation}
We use Mistral-7B as our base pre-trained model. The supervised fine-tuning and preference learning experiments are conducted with a maximum of $4 \times 40$GB GPUs (NVIDIA A100). 

We choose the learning rates 5e-6 and 1e-6 for SFT and DPO training, respectively, with a cosine learning rate scheduler. The maximum sequence length of models is $512$. We train the model with a batch size of $128$ and $32$ for SFT and DPO, respectively. For DPO, we follow the DPO paper to set the KL constraint parameter $\beta$ as $0.1$. Each sample in DPO is a set of step-level preference data decomposed by MCTS. We set the max length for each step as $64$. The number of MCTS iterations is set as $K=5$ for all tasks.

For arithmetic reasoning, we combine the problems in GSM8K and MATH training sets as the prompt data containing a total of $24$K samples for preference learning. For each sample, we conduct MCTS with an initial breadth of $b_1=5$ and decrease it to $b_2=3$ for the subsequent steps, with a maximum search depth $d=4$. It takes about $2$ minutes per sample to collect the step-level preferences via MCTS. This requires about 30 A100 days of compute to train one whole epoch. In practice, we can adopt an early stop when the performance saturates, which usually only needs $30\%$ of the training data.

For commonsense reasoning, we combine the training data of ARC, AI2Science, OBQA, and CSQA, which produces a total of $12$K samples. As the model generations are more diversified on these tasks, we set the initial breadth as $b_1=4$ and decrease it to $b_2=2$ for subsequent steps. As the intermediate reasoning chains are relatively shorter than those in arithmetic reasoning, we set the maximum search depth $d=3$. Likewise, we also adopt an early stop at around $50\%$ of the training progress where the performance saturates.

\paragraph{Hyperparameter Tuning of MCTS.}
We compare the performance in commonsense reasoning when employing different searching breadths in MCTS. Table~\ref{tab:hyp-mcts} shows how different search heuristics impact learning performance. O2 produces better performance, highlighting the importance of increasing the search space at the beginning point of MCTS. One can efficiently reduce compute while maintaining good performance by using a small search space for the subsequent steps. For future work, we will explore the hyperparameter settings in MCTS, including the search breadth, depth, number of steps, and iteration time, to probe the cost--performance tradeoff of our MCTS-enhanced iterative learning framework.
\begin{table}[h]
\begin{center}
\begin{small}
\begin{tabular}{lccccccc}
\toprule
Approach & ARC-e & ARC-c & AI2Sci-e & AI2Sci-m & OBQA & CSQA & SciQ \\
\midrule
SFT Baseline & $69.2$ & $60.6$ & $74.9$ & $70.9$ & $59.8$ & $54.1$ & $80.8$ \\
\midrule
O1 ($b_1=3, b_2=3$) & $88.4$ & $74.7$ & $92.1$ & $88.5$ & $77.8$ & $73.2$ & $88.3$ \\
O2 ($b_1=4, b_2=2$) & $88.5$ & $76.4$ & $91.7$ & $88.2$ & $79.2$ & $74.8$ & $88.5$ \\
\bottomrule
\end{tabular}
\end{small}
\end{center}
\caption{Result comparison of using different search breadths in MCTS. For O2, we have a broader spectrum for the initial step and narrow the search space for the subsequent steps of each path.}
\label{tab:hyp-mcts}
\vskip -0.1in
\end{table}

\paragraph{Prompt Example.}
See an example of the evaluation prompt we use for self-evalution in Table~\ref{tab:eval-prompt}. For more details, please refer to our implementation code.
\begin{table}[h]
\centering
\small
\caption{Evaluation Prompt Template. The text underlined will be replaced with content from different examples.}
\label{tab:eval-prompt}
\begin{tabular}{l}
\toprule
\textbf{QUESTION:} \underline{Which of the following is an example of the formation of a mixture?} \\
\underline{Answer Choices: (A) rust forming on an iron nail (B) sugar crystals dissolving in water (C) sodium and} \\ \underline{chlorine forming table salt (D) hydrogen and oxygen reacting to produce water} \\
\\
\textbf{EXAMPLE ANSWER:} \underline{The answer is (B) sugar crystals dissolving in water} \\
\\
\textbf{PROPOSED SOLUTION:} \underline{The formation of a mixture occurs when two or more substances are} \\
\underline{combined together without changing their individual properties. In the given options, rust forming on an} \\
\underline{iron nail is an example of the formation of a mixture. The iron nail and the oxygen in the air combine to} \\
\underline{form iron oxide, which is a mixture. The answer is A.} \\
\\
\textbf{QUESTION}: Evaluate if the proposed solution is logically heading in the correct direction. \\
Provide an answer of (A) correct or (B) incorrect. \\
\\
\textbf{ANSWER}: The answer is \\
\bottomrule
\end{tabular}
\vskip -0.1in
\end{table}

\section{Further Analysis}\label{app:analysis}
\paragraph{Reward Criteria in MCTS.}
We probe the effect of different reward guidance of MCTS in terms of both searching and training. Table~\ref{tab:mcts} shows how different reward signals impact the pass rate of searching. The guidance of outcome correctness is substantially dominant in eliciting correct outcomes. We see that MCTS can produce significant improvement across various tasks with the reward signals integrated of outcome correctness and self-evaluation, increasing the baseline performance from $60.6\%$ to $83.0\%$ on ARC-C, $70.9\%$ to $90.5\%$ on AI2Sci-M, and $75.9\%$ to $85.8\%$ on GSM8K. We observe a significant performance gain from learning when using greedy decoding on commonsense reasoning. For example, learning increases the accuracy to $76.4\%$ (+$16.4\%$) on ARC-C, compared to the increase of $9.1\%$ on MCTS performance. This suggests a substantial improvement in the model's policy when applying our MCTS-enhanced iterative learning to tasks that the initial policy is not good at. Furthermore, the ablation study on the reward components shows consistent improvement brought by self-evaluation to increase the MCTS performance in both before- and after- learning cases, suggesting the effectiveness of the integration of self-evaluation in our approach.
\begin{table}[h]
    \centering
    \caption{Pass Rates when Ablating MCTS Settings. SE represents the guidance from self-evaluation.}
    \label{tab:mcts}
    \begin{small}
    \begin{tabular}{lccccc}
        \toprule
         Decoding Strategy & After Learning & ARC-C & AI2Sci-M & GSM8K \\
         \midrule
         \multirow{2}{*}{Greedy Decoding} & \ding{55} & $60.6$ & $70.9$ & $75.9$ \\
          & $\checkmark$ & \deltavalup{76.4}{16.4} & \deltavalup{88.2}{17.3} & \deltavalup{80.7}{5.2} \\
         \midrule
         \multirow{2}{*}{MCTS w/o SE} & \ding{55} & $82.5$ & $87.3$ & $84.4$ \\
          & $\checkmark$ & \deltavalup{91.0}{8.5} & \deltavalup{96.1}{9.8} & \deltavalup{89.0}{5.6} \\
         \midrule
         \multirow{2}{*}{MCTS} & \ding{55} & $83.0$ & $90.5$ & $85.8$ \\
         & $\checkmark$ & \deltavalup{92.1}{9.1} & \deltavalup{97.3}{6.8} & \deltavalup{90.2}{4.4} \\
        \bottomrule
    \end{tabular}
    \end{small}
\end{table}

\begin{figure}[t]
    \centering
    \includegraphics[width=.95\textwidth]{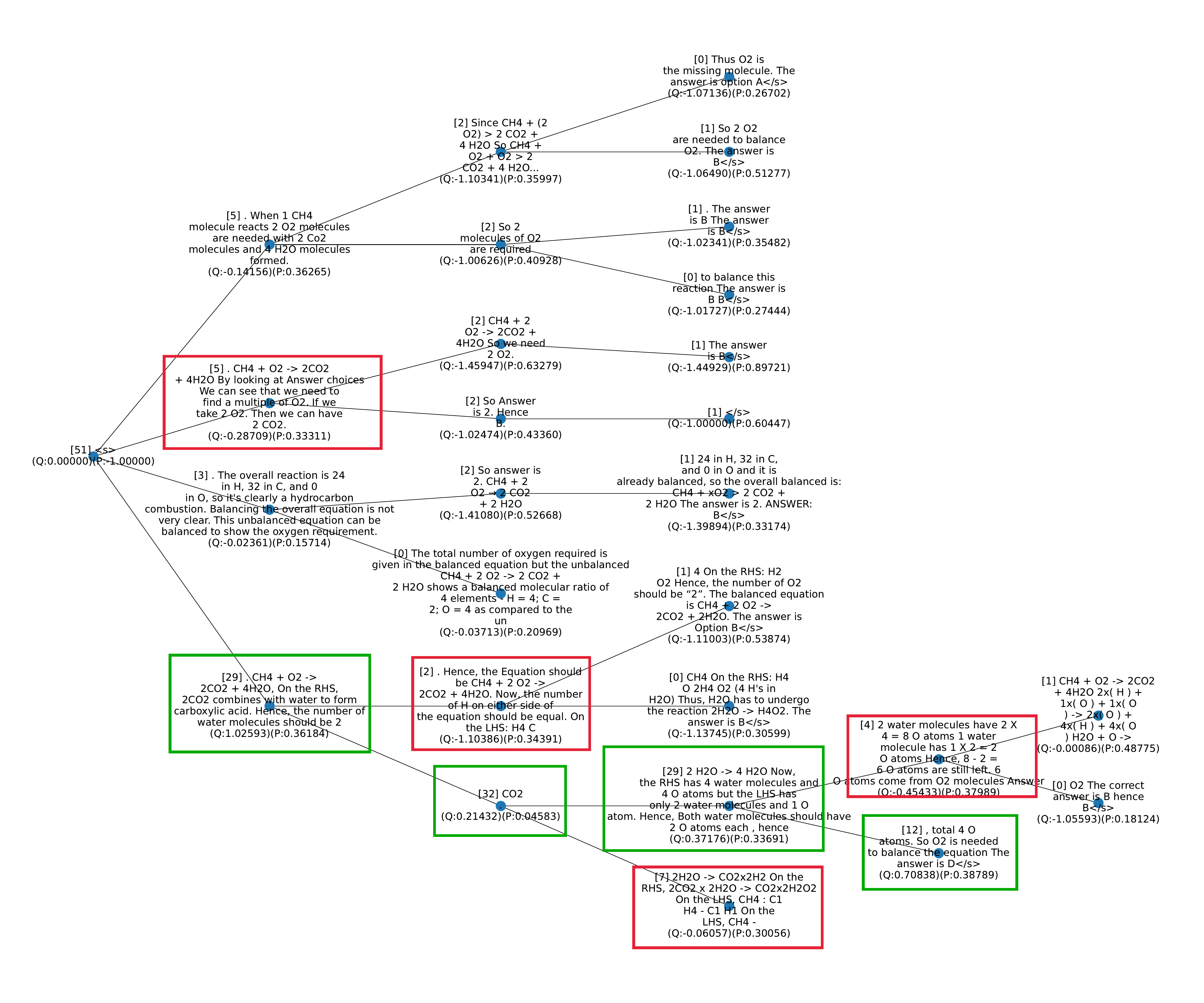}
    \caption{Example of the result search tree of a science question ``An unbalanced equation for the reaction of methane gas ($\mathrm{CH}_{4}$) with oxygen is shown below. $\mathrm{CH}_{4} + \Box \mathrm{O}_{2} \rightarrow 2\mathrm{CO}_{2} + 4\mathrm{H}_{2}\mathrm{O}$ How many molecules of oxygen gas ($\mathrm{O}_{2}$) are needed to properly balance this equation? Answer Choices: (A) 1 (B) 2 (C) 3 (D) 4''. The ground truth answer is ``(D) 4''. Here, we set the search breadth as $b_1=4, b_2=2$. The numbers at the beginning of each sequence indicate the visit count $N$ of the corresponding node, while the $Q$ and $P$ values at the end of the sequence represent the $Q$ values and the sentence probability, respectively.}
    \label{fig:sqa-4x2}
\end{figure}

\begin{figure}
    \centering
    \includegraphics[width=.95\textwidth]{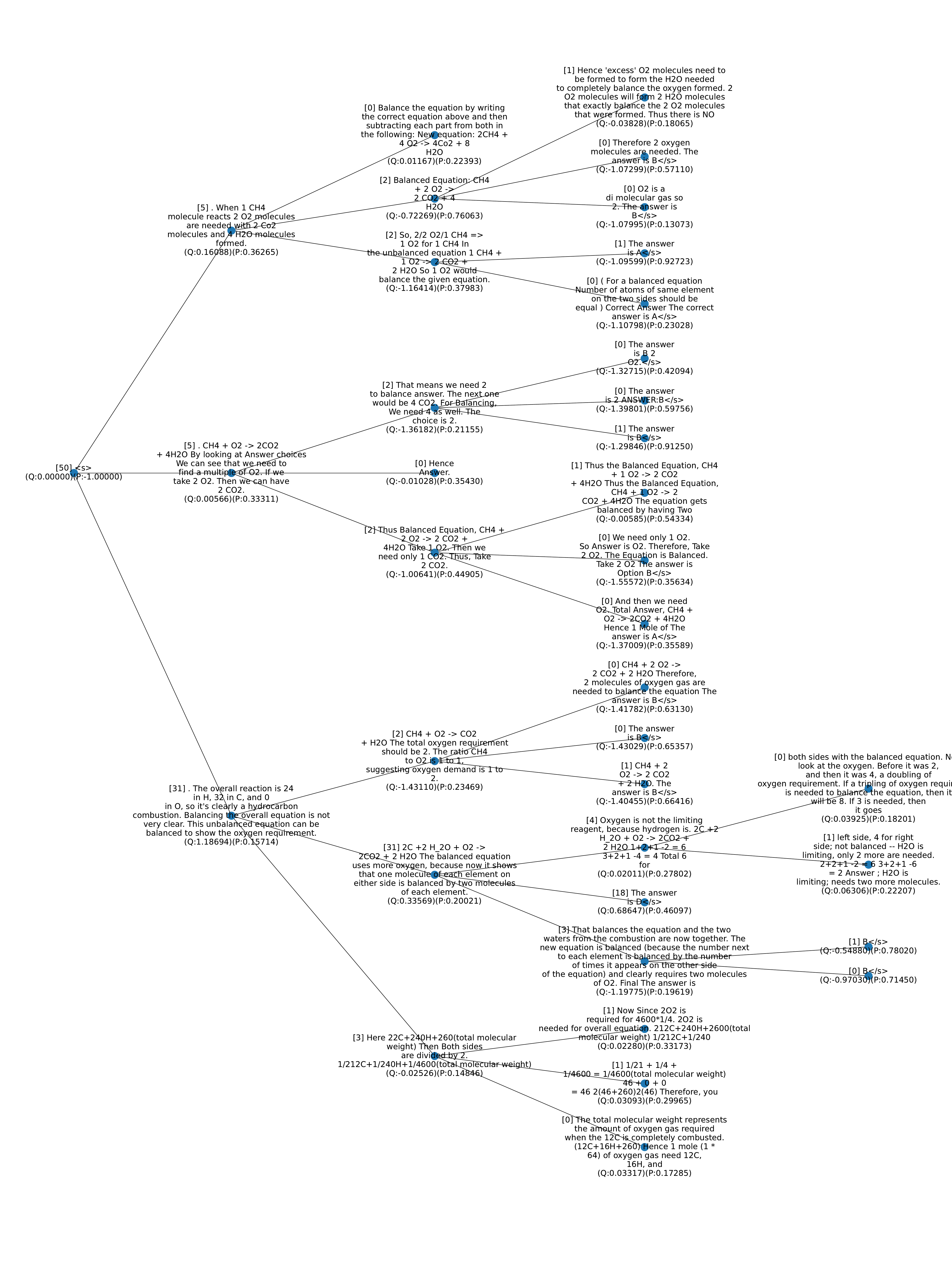}
    \caption{Example of the result search tree of the same science question as in Figure~\ref{fig:sqa-4x2}. Here, we set the search breadth as $b_1=3, b_2=3$.}
    \label{fig:sqa-3x3}
\end{figure}

\begin{figure}
    \centering
    \includegraphics[width=.95\textwidth]{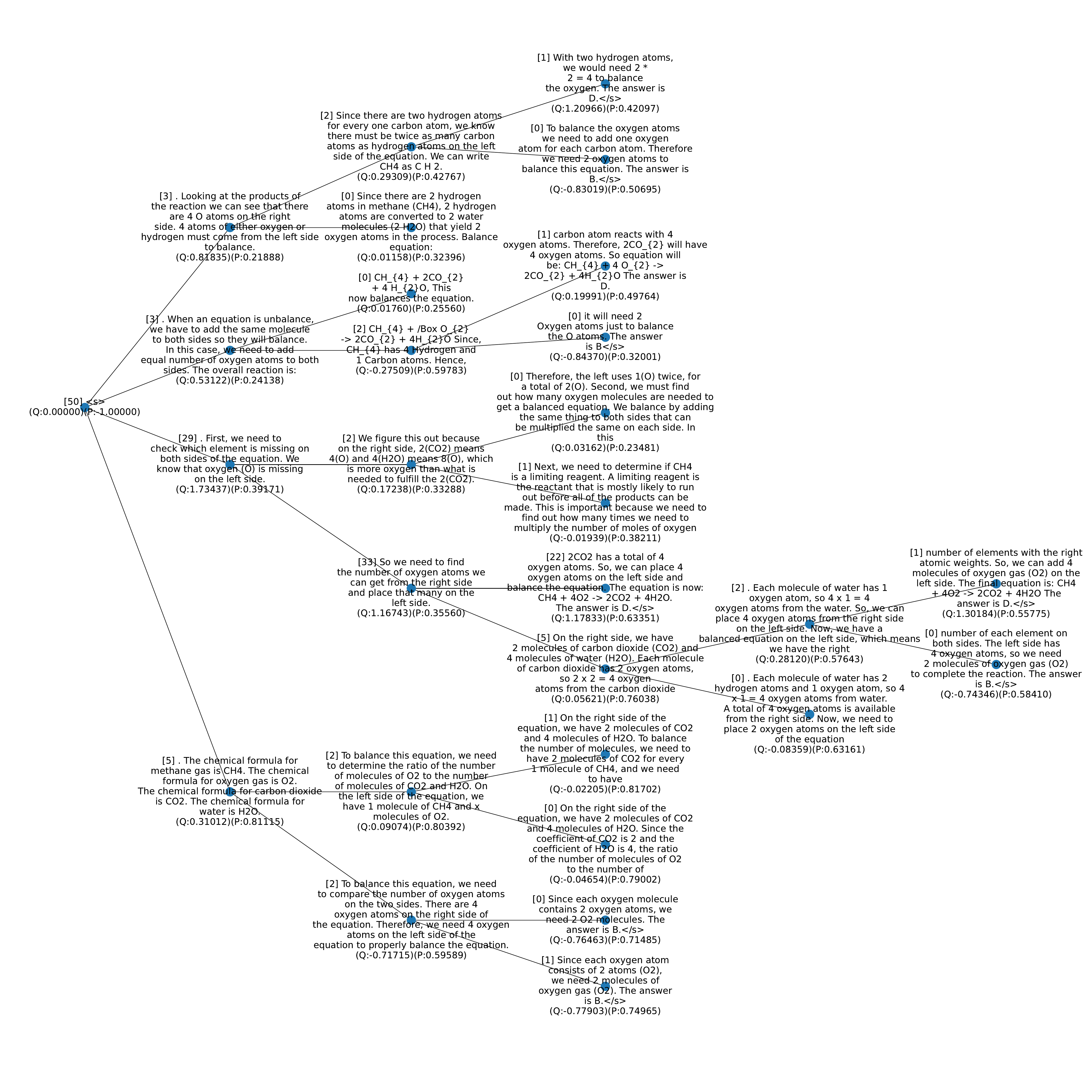}
    \caption{Example of the result search tree of the same science question as in Figure~\ref{fig:sqa-4x2}. Here, we use the policy after preference learning and set the search breadth as $b_1=4, b_2=2$.}
    \label{fig:sqa-4x2-train}
\end{figure}

\begin{figure}
    \centering
    \includegraphics[width=.9\textwidth]{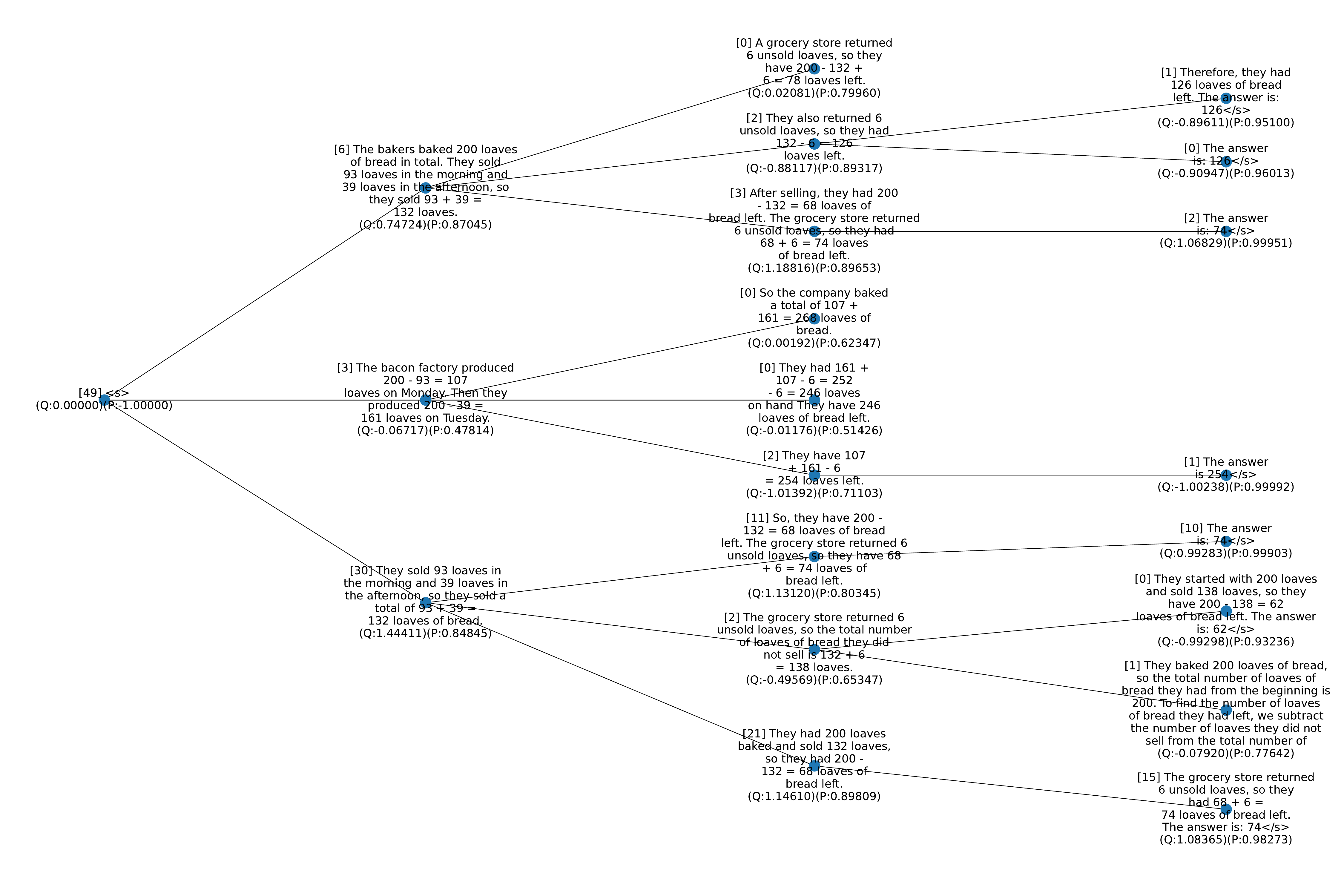}
    \caption{Example of the result search tree of a GSM8K question ``The bakers at the Beverly Hills Bakery baked 200 loaves of bread on Monday morning. They sold 93 loaves in the morning and 39 loaves in the afternoon. A grocery store returned 6 unsold loaves. How many loaves of bread did they have left?''. The example solution is ``The Bakery sold 93 + 39 = 132 loaves. The Bakery made 200 loaves and sold 132, leaving 200 - 132 = 68 loaves remaining. The grocery store returned 6 loaves, so there were 6 + 68 = 74 loaves left.''. The policy we use here is the one only tuned for $1$ epoch on SFT training data. We conduct MCTS with breadth $b_1=5, b_2=3$. Duplicate generations are merged into one node.}
    \label{fig:gsm-e1}
\end{figure}

\begin{figure}
    \centering
    \includegraphics[width=.9\textwidth]{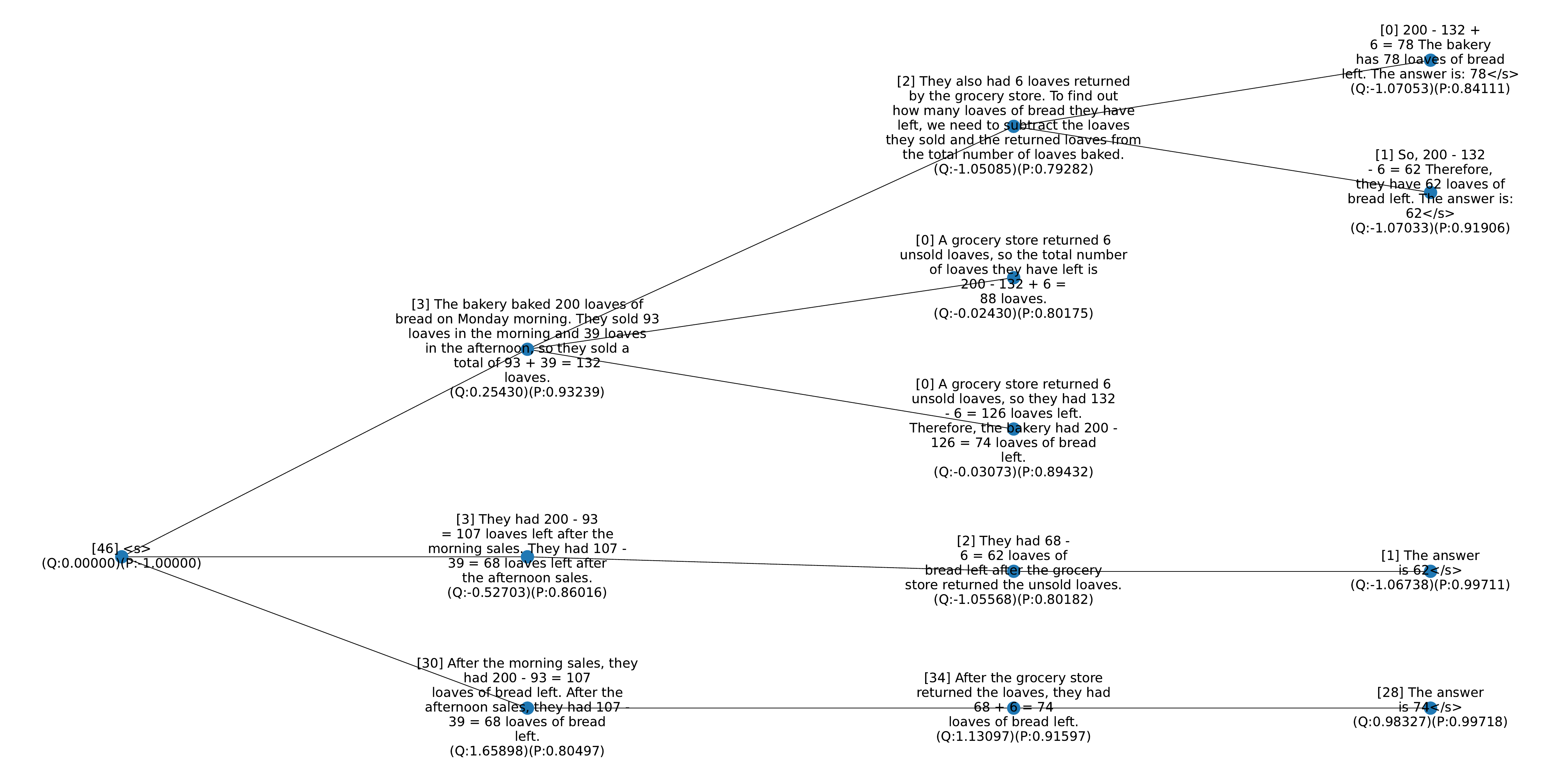}
    \caption{Example of the result search tree of the same GSM8K question as in Figure~\ref{fig:gsm-e1} with the same search breadth. We use the policy tuned after $3$ epochs to sample the generations.}
    \label{fig:gsm-e3}
\end{figure}

\paragraph{Qualitative Analysis on Collected Preferences.}
We show examples of the result search trees elicited via MCTS on different tasks in Figures~\ref{fig:sqa-4x2}--\ref{fig:gsm-e3}. 

Figures~\ref{fig:sqa-4x2} and ~\ref{fig:sqa-3x3} show the result search trees to answer the same science question using MCTS employed with different search breadths. We see that MCTS not only figures out the correct answer (\textit{i.e.}, the option ``D'') via broad searching but also serves as a policy improvement optimizer to collect steps along this path as positive samples for preference learning. For example, the $Q$ values of the preference pair at the last step (at the bottom right of Figure~\ref{fig:sqa-4x2}) are $0.70838$ and $-0.45433$, compared to the original probability in the policy generation as $0.37989$ and $0.38789$. Compared to searching with breadth $b_1=4, b_2=2$ in Figure~\ref{fig:sqa-4x2}, Figure~\ref{fig:sqa-3x3} shows that a higher breadth for the subsequent steps can produce an even larger search tree. However, as we only collect preference pairs alongside the paths leading to correct prediction, these two search heuristics can result in preference data of similar size. 

Figure~\ref{fig:sqa-4x2-train} shows the search tree using the trained policy on commonsense reasoning. Compared to the one generated by the initial policy in Figure~\ref{fig:sqa-4x2}, the policy has a higher chance to elicit correct reasoning chains, as we see more successful predictions of the ground-truth option ``D''. We also observe that the policy tends to generate longer reasoning chains after being motivated to conduct chain-of-thought reasoning with fine-grained process supervision. 

On arithmetic reasoning, we also probe the impact of diversity in model generations using policies trained for different numbers of epochs in SFT. 
Figures~\ref{fig:gsm-e1} and \ref{fig:gsm-e3} show the elicited search trees with data sampled by policies corresponding to different levels of diversity, where the policy used in Figure~\ref{fig:gsm-e1} has generations with higher diversity. With higher diversity, MCTS can explore more alternatives of the correct solutions, as there are more paths of correct predictions, as shown in Figure~\ref{fig:gsm-e1} than Figure~\ref{fig:gsm-e3}. Furthermore, higher diversity with reasonable quality also provide more fine-grained supervision signals as there are more branches alongside the reasoning path of correct predictions.


\section{Extended Experiments}\label{sec:ext-exp}
\paragraph{Loss Function.}
DPO is one of the reward-model-free loss functions we can use for preference learning. We now illustrate the generalizability of our approach using another loss function, Identity Preference Optimization (IPO)~\citep{DBLP:journals/corr/abs-2310-12036}, which addresses the overfitting problem of DPO. Table~\ref{tab:loss} shows that IPO achieves similar performance as DPO. In practice, we find that IPO boosts the reasoning on validation tasks while maintaining a more stable performance on the held-out dataset, as indicated by the higher accuracy $89.8\%$ obtained on SciQ. 
\begin{table}[h]
\begin{center}
\begin{small}
\caption{Result comparison of employing our approach with different loss functions.}
\label{tab:loss}
\begin{tabular}{lccccccc}
\toprule
Approach & ARC-e & ARC-c & AI2Sci-e & AI2Sci-m & OBQA & CSQA & SciQ \\
\midrule
SFT Baseline & $69.2$ & $60.6$ & $74.9$ & $70.9$ & $59.8$ & $54.1$ & $80.8$ \\
\midrule
O1 (IPO) & $88.1$ & $75.1$ & $92.1$ & $89.6$ & $76.8$ & $74.3$ & $89.8$ \\
O2 (DPO) & $88.5$ & $76.4$ & $91.7$ & $88.2$ & $79.2$ & $74.8$ & $88.5$ \\
\bottomrule
\end{tabular}
\end{small}
\end{center}
\vskip -0.1in
\end{table}

\paragraph{Base Model.}
We extensively validate the generalizability of our approach on Llama2-13B~\citep{DBLP:journals/corr/abs-2307-09288} on arithmetic reasoning. We employ the same process of SFT on Arithmo and preference learning with DPO on GSM8K and MATH. This experiment is done on a maximum of $2 \times 80$GB GPUs (NVIDIA A100). 

\begin{table}[h]
\begin{center}
\begin{small}
\caption{Result comparison (accuracy $\%$) for Llama2-13B on arithmetic tasks. }
\label{tab:math-llama}
\begin{tabular}{lccccc}
\toprule
Approach & Base Model & GSM8K-CoT & GSM8K-PoT & MATH-CoT \\

\midrule
SFT (Arithmo) & \multirow{2}{*}{Llama2-13B} & $74.5$ & $62.3$ & $23.8$ \\
Ours & & \deltavalup{78.9}{4.4} & \deltavalup{67.0}{4.7} & \deltavalup{26.1}{2.3} \\
\bottomrule
\end{tabular}
\end{small}
\end{center}
\vskip -0.1in
\end{table}

\newpage

\end{document}